\documentclass[11pt]{article} 
\usepackage[english]{babel}
\usepackage[a4paper,margin=1in]{geometry} 
\usepackage{authblk} 
\usepackage[colorlinks=true, linkcolor=black, citecolor=black, urlcolor=blue, filecolor=blue,breaklinks=true]{hyperref} 
\usepackage{setspace} 
\usepackage{lineno} 
\usepackage{csquotes} 
\usepackage{ragged2e}
\usepackage{array} 

\usepackage[defaultmathsizes]{mathastext}
\usepackage{amsmath}

\usepackage[capitalise, noabbrev, nameinlink]{cleveref}
\usepackage[labelsep=period]{caption} 
\crefdefaultlabelformat{#2#1#3}
\Crefname{figure}{Figure}{Figures}
\Crefname{table}{Table}{Tables}
\Crefname{section}{\textbf{Section}}{\textbf{Sections}}

\newcommand{\best}[1]{{\textbf{#1}}}
\newcommand\var[1]{\scriptsize$\pm$#1}

\newcommand{\ie}{\emph{i.e., }}
\newcommand{\eg}{\emph{e.g., }}

\newcommand{\wrt}{\emph{w.r.t. }}

\newcommand{\aka}{\emph{aka. }}

\usepackage{tikz}
\usetikzlibrary{positioning,calc, matrix, 3d, positioning, fit, arrows, shadings, fadings, shadows}
\usetikzlibrary{tikzmark,calc}
\usepackage{pgfplots}
\usetikzlibrary{arrows.meta}
\usetikzlibrary{shadows.blur}
\usetikzlibrary{shapes.symbols}
\usetikzlibrary{backgrounds}
\usepgfplotslibrary{colormaps}

\usepackage{graphicx}
\graphicspath{{./main_figs/}{./supplement/suppl_figs/}} 
\DeclareGraphicsExtensions{.pdf,.jpeg,.JPG,.png,.PNG, .eps, .tiff}
\usepackage{subcaption} 
\DeclareCaptionLabelFormat{bold}{{(#2)}} 
\usepackage{pgffor} 
\usepackage{alphalph} 
\usepackage[labelfont=bf, textfont=bf, singlelinecheck=off, textfont=footnotesize]{caption} 
\usepackage{booktabs}
\usepackage{multirow}
\usepackage{rotating}

\usepackage{amsmath,amsfonts,bm}

\def\eqref#1{equation~\ref{#1}}
\def\Eqref#1{Equation~\ref{#1}}

\def\1{\bm{1}}

\def\vc{{\bm{c}}}

\def\vh{{\bm{h}}}

\def\vx{{\bm{x}}}

\def\mI{{\bm{I}}}

\def\mO{{\bm{O}}}

\def\mR{{\bm{R}}}

\DeclareMathAlphabet{\mathsfit}{\encodingdefault}{\sfdefault}{m}{sl}
\SetMathAlphabet{\mathsfit}{bold}{\encodingdefault}{\sfdefault}{bx}{n}

\def\gG{{\mathcal{G}}}

\def\sA{{\mathbb{A}}}

\def\sD{{\mathbb{D}}}

\def\sG{{\mathbb{G}}}

\def\sR{{\mathbb{R}}}

\newcommand{\KL}{D_{\mathrm{KL}}}

\usepackage{xurl}
\usepackage{natbib}

\usepackage{xcolor}
\usepackage{caption}

\definecolor{sectioncolor}{HTML}{9c4847}
\definecolor{black}{HTML}{000000}
\definecolor{linkcolor}{HTML}{008ebe}
\definecolor{revisecolor}{HTML}{C33636}

\usepackage{titlesec}
\titleformat{\section}
{\color{sectioncolor}\normalfont\Large\bfseries}
{\thesection}{1em}{}

\titleformat{\subsection}
{\color{sectioncolor}\normalfont\large\bfseries}
{\thesubsection}{1em}{}

\DeclareCaptionFont{figuretitlecolor}{\color{sectioncolor}}
\DeclareCaptionFont{figurecaptioncolor}{\color{black}}
\DeclareCaptionFont{tabletitlecolor}{\color{sectioncolor}}
\DeclareCaptionFont{tablecaptioncolor}{\color{black}\footnotesize}
\hypersetup{
    colorlinks=true,
    linkcolor=linkcolor,
    urlcolor=linkcolor,
    citecolor=linkcolor
}

\makeatletter
\renewcommand\maketitle{\par
  \begingroup
    \flushleft
    \Large{\@title}\par
  \endgroup
  \vspace{0.5em}
  \begin{flushleft}
    \@author
  \end{flushleft}
}
\makeatother

\renewenvironment{abstract}{
  \begin{flushleft}
    \textbf{\abstractname}
  \end{flushleft}
  \vspace{-1.5em}
  \par
  \noindent\justify
  \parfillskip=0pt
}{
  \par
}

\newif\ifmain
\maintrue 

\ifmain
    \title{Text-guided Diffusion Model for 3D Molecule Generation}
\else
    \title{Supplementary Information of Text-guided Diffusion Model for 3D Molecule Generation}
\fi

\author[1,3]{Yanchen Luo}
\author[1,\#]{Junfeng Fang}
\author[1]{Sihang Li}
\author[2]{Zhiyuan Liu}
\author[1]{Jiancan Wu}
\author[2]{An Zhang}
\author[1,*]{Wenjie Du}
\author[1,*]{Xiang Wang}
\affil[1]{University of Science and Technology of China, Hefei, Anhui, China}
\affil[2]{National University of Singapore, Singapore}
\affil[3]{Lead contact: \href{luoyanchen@mail.ustc.edu.cn}{luoyanchen@mail.ustc.edu.cn}} 
\affil[$\#$]{Equal contribution}
\affil[*]{Correspondence: \href{duwenjie@mail.ustc.edu.cn}{duwenjie@mail.ustc.edu.cn}, \href{xiangwang1223@gmail.com}{xiangwang1223@gmail.com}}
\affil[ ]{ } 
\date{} 

\newcommand{\makeAbstract}{
\renewcommand{\abstractname}{\color{sectioncolor}SUMMARY}
\begin{abstract}
\noindent \color{linkcolor}The \textit{de novo} generation of molecules with targeted properties is crucial in biology, chemistry, and drug discovery.
Current generative models are limited to using single property values as conditions, struggling with complex customizations described in detailed human language.
To address this, we propose the text guidance instead, and introduce TextSMOG, a new \textit{\underline{Text}-guided \underline{S}mall \underline{Mo}lecule \underline{G}eneration Approach via 3D Diffusion Model} which integrates language and diffusion models for text-guided small molecule generation. This method uses textual conditions to guide molecule generation, enhancing both stability and diversity. Experimental results show TextSMOG's proficiency in capturing and utilizing information from textual descriptions, making it a powerful tool for generating 3D molecular structures in response to complex textual customizations.
\\
\end{abstract}
\subsection{Keywords}
{\color{linkcolor}Small molecule generation, geometry generation, Diffusion model}
}
\begin{document}
    \setstretch{1.15} 

    \ifmain
        \maketitle
    	\makeAbstract
    	\clearpage
		\section{INTRODUCTION}
\textit{De novo} molecule design, the process of generating molecules with specific, chemically viable structures for target properties, is a cornerstone in the fields of biology, chemistry, and drug discovery \citep{Hajduk_Greer_2007, Mandal_Moudgil_Mandal_2009, Pyzer-Knapp_2015, Khaled_H_Barakat_2014}.
It not only allows for the creation of subject molecules but also provides insights into the relationship between molecular structure and function, enabling the prediction and manipulation of biological activity.
Constrained by the immense diversity of chemical space, manually generating property-specific molecules remains a daunting challenge \citep{GML}.
However, the generation of molecules that precisely meet specific requirements, including the creation of tailor-made molecules, is a complex task due to the vastness of the chemical space and the intricate relationship between molecular structure and function.
Overcoming this challenge is crucial for advancing our understanding of biological systems and for the development of new therapeutic agents.
In recent years, machine and deep learning methods have initiated a paradigm shift in the molecule generation \citep{Alcalde_Ferrer_Plou_2007, anand2022protein, CVGAE, MoFlow, E-NFs, G-SchNet, MolCA, E-GCL, SimSGT}, which enable the direct design of 3D molecular geometric structures with the desired properties  \citep{MDM, DGSM, CVGAE}.
Notably, diffusion models \citep{Diffusion, DDPM}, specifically equivariant diffusion models \citep{EDM, EEGSDE}, have gradually enter the center of the stage with its outstanding performance.
The core of this method is to introduce diffusion noise on molecular data, and then learn a reverse process in either \textit{unconditional} or \textit{conditional} manners to denoise this corruption, thereby crafting desired 3D molecular geometries.
Meanwhile, some conditional inputs (\eg polarizability $\alpha = 100 \textnormal{ Bohr}^3$) could be applied for constraining the model to generate more specific molecules types.

However, despite the promise of these methods, a significant proportion of molecules generated by diffusion models do not meet the practical needs of researchers.
For instance, they may lack the desired biological activity, exhibit poor pharmacokinetic properties, or be synthetically infeasible.
This would be due to the fact that, on one hand, searching for suitable molecules in drug design typically requires consideration of multiple properties of interest (e.g., simultaneously characterized by specific polarizability, orbital energy, properties like aromaticity, and distinct functional groups) \citep{Honório_Moda_Andricopulo_2013, cG-SchNet, MGCVAE}.
On the other hand, humans seem to struggle with conveying their needs precisely to the model.
While a text segment such as ''This molecule is an aromatic compound, with small HOMO-LUMO gaps and possessing at least one carboxyl group'' can accurately describe human requirements and facilitate communication among humans, it is still challenging to directly convey this 'thoughts' to the model.
Therefore, we aspire to develop a method that allows for the interactive inverse design of 3D molecular structures through natural language.
In other words, we aim to create a system where researchers can describe the properties they want in a molecule using natural language, and the system will generate a molecule that meets these requirements.
This aspiration prompts us to explore text guidance in diffusion models, emphasizing the necessity for models adept at precise language understanding and molecule generation.

Towards this end, we propose TextSMOG, a new \underline{text}-guided \underline{s}mall \underline{mo}lecule \underline{g}eneration approach.
The basic idea is to combine the capabilities of the advanced language models \citep{BERT, RoBERTa, SciBERT, T5, GPT-3, GPT-4, ReactXT, 3DMoLM} with high-fidelity diffusion models, enabling a sophisticated understanding of textual prompts and accurate translation into 3D molecular structures.
TextSMOG accomplishes this through integrating textual information with a conversion module that conditions a pre-trained equivariant diffusion model (EDM) \citep{EDM}, following the multi-modal fusion fashion \citep{MoMu, MoFlow, Text2Mol, MolT5, MolCA, ProtT3, MolTC}.
Specifically, at each denoising step, TextSMOG first generates reference geometry, an intermediate conformation that encapsulates the textual condition signal, through a multi-modal conversion module.
Equipped with language and molecular encoder-decoder, corresponding to the textual condition.
Then the reference geometry guides the denoising of each atom within the pre-trained unconditional EDM, gradually modifying the molecular geometry to match the condition while maintaining chemical validity.
By incorporating valuable language knowledge into the pre-trained diffusion model, TextSMOG enhances the generation of valid and stable 3D molecular conformations that align with a spectrum of diverse directives.
This is achieved without the need for exhaustive training on each specific condition, demonstrating the model's ability to generalize from the language input.
This integration allows for the incorporation of valuable language knowledge in the high-fidelity pre-trained diffusion model, thereby enabling the conditional generation contingent upon a spectrum of diverse directives, while enhancing the generation of valid and stable 3D molecular conformations , without specific exhaustive training of the condition

We applied TextSMOG to the standard quantum chemistry dataset QM9 \citep{QM9} and a real-world text-molecule dataset from PubChem \citep{PubChem}.
The experimental results show that TextSMOG accurately captures single or multiple desired properties from textual descriptions, thereby aligning the generated molecules with the desired structures.
Notably, TextSMOG outperforms leading diffusion-based molecule generation baselines (\eg EDM \citep{EDM}, EEGSDE \citep{EEGSDE}) in terms of both the stability and diversity of the generated molecules.
This is evidenced by higher scores on metrics such as the Tanimoto similarity to the target structure, the synthetic accessibility of the generated molecules, and the diversity of the generated molecule set.
Furthermore, when applied to real-world textual excerpts, TextSMOG demonstrates its generative capability under general textual conditions.
These findings suggest that TextSMOG constitutes a versatile and efficient text-guided molecular diffusion framework.
As an advanced intelligent agent, it can effectively comprehend the meaning of textual commands and accomplish generation tasks, thereby paving the way for a more in-depth exploration of the molecular space.
		\section{RESULTS}
In this section, we present the architecture and the experimental results of our proposed TextSMOG model, showcasing its ability to generate molecules with desired properties.

\subsection{Architecture}
To evaluate our model, we employ the QM9 dataset \citep{QM9}, which is a standard benchmark containing quantum properties and atom coordinates of over 130K molecules, each with up to 9 heavy atoms (C, N, O, F).
For the purpose of training our model under the condition of textual descriptions, we have curated a subset of molecules from QM9 and associated them with real-world descriptions.
These descriptions are sourced from PubChem \citep{PubChem}, one of the most comprehensive databases for molecular descriptions, and are linked to the molecules in QM9 based on their unique SMILES.

PubChem aggregates extensive annotations from a diverse array of sources, such as ChEBI \citep{ChEBI}, LOTUS \citep{LOTUS}, and T3DB \citep{T3DB}.
Each of these sources offers an emphasis on the physical, chemical, or structural attributes of molecules.
Additionally, we have employed a set of textual templates to generate corresponding descriptions based on the quantum properties of the molecules, thereby enriching the content of the dataset and supplementing textual context for those molecules lacking real-world descriptions.
This process has enriched QM9 into a dataset of chemical molecule-textual description pairs.
Our proposed TextSMOG model, illustrated in Figure \ref{figure-framework}, is built upon the pre-trained unconditional diffusion model EDM \citep{EDM}. It integrates the textual information into the conditional signal of diffusion models by employing a reference geometry that is updated at each step based on the textual prompt. The final molecular geometry is generated by gradually denoising an initial geometry, while noise is added at each step during the forward process until the molecular geometry is fully noise-corrupted.

\begin{figure}[t]
  \centering
  \includegraphics[]{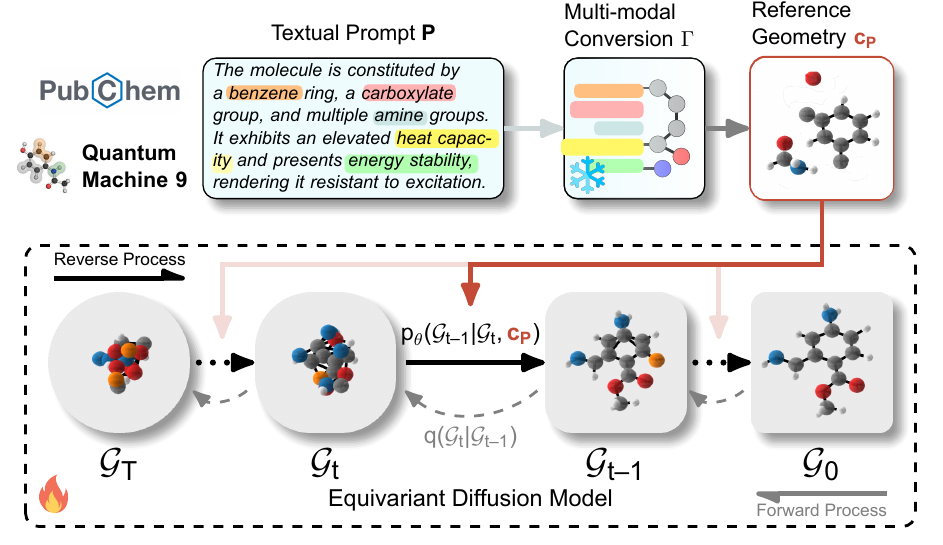}
  \caption{Architecture of Our Text-guided Small Molecule Generation via Diffusion Model (TextSMOG). The model starts with an initial geometry ($\gG_{T}$) and gradually denoises it to generate the final molecular geometry. The reference geometry ($\vc_{\textbf{P}}$), updated at each step based on the textual prompt ($\textbf{P}$), is employed to integrate the textual information into the conditional signal of diffusion models. Flame \includegraphics[width=0.02\textwidth]{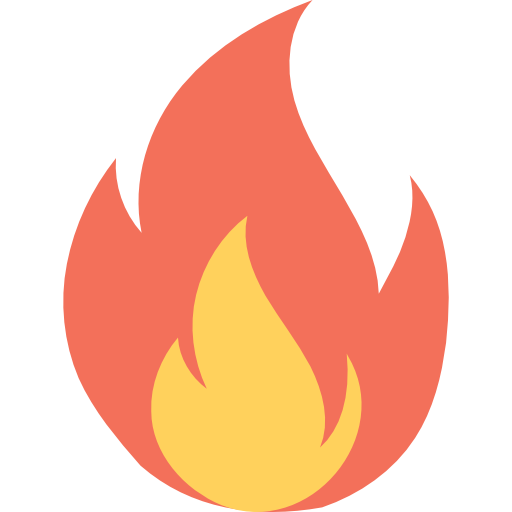} denotes tunable modules, while snowflake \includegraphics[width=0.02\textwidth]{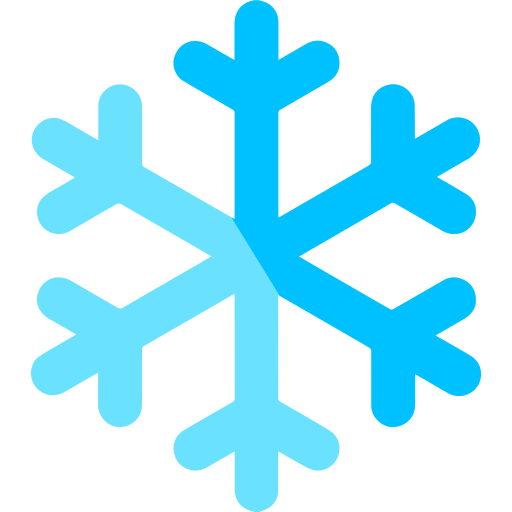} indicates frozen modules.}
  \label{figure-framework}
\end{figure}

\subsection{Experiment on Single Quantum Properties Conditioning}\label{subsection-exp_single}
Following EDM \citep{EDM}, we first evaluate our TextSMOG on the task of generating molecule conditioning on a single desired quantum property in QM9.
Then we compare our TextSMOG with several baselines to demonstrate the effectiveness of our model on single quantum properties conditioning molecule generation.

\textbf{Setup.}
We follow the same data preprocessing and partitions as in EDM \citep{EDM}, which results in 100K/18K/13K molecule samples for training/validation/test respectively.
In order to assess the quality of the conditional generated molecules \wrt to the desired properties, we use the property classifier network $\phi_{p}$ introduced by \citep{EGNN}.
Then for the impartiality, the training partition is further split into two non-overlapping halves $\sD_{a}$ and $\sD_{b}$ of 50K molecule samples each.
The property classifier network $\phi_{p}$ is trained on the first half $\sD_{a}$, while our TextSMOG is trained on the second half $\sD_{b}$.
This ensures that there is no information leak and the property classifier network $\phi_{p}$ is not biased towards the generated molecules from TextSMOG.
Then $\phi_{p}$ is evaluated on the generated molecule samples from TextSMOG as we introduce in the following.

\textbf{Metrics.}
Following \citep{EDM}, we use the mean absolute error (MAE) between the properties of generated molecules and the ground truth as a metric to evaluate how the generated molecules align with the condition (see the supplementary information for details).
We generate 10K molecule samples for the evaluation of $\phi_{p}$, following the same protocol as in EDM.
Additionally, we then measure novelty \citep{GraphVAE}, atom stability \citep{EDM}, and molecule stability \citep{EDM} to demonstrate the fundamental molecule generation capacity of the model (also see the supplementary information for details).

\textbf{Baseline.}
We compare our TextSMOG with a direct baseline conditional EDM \citep{EDM} and a recent work EEGSDE which takes energy as guidance \citep{EEGSDE}.
We also compare two additional baselines ``U-bound'' and ``\texttt{\#}Atoms'' introduced by \citep{EDM}.
In the ``U-bound'' baseline, any relation between molecule and property is ignored, and the property classifier network $\phi_{p}$ is evaluated on $\sD_{b}$ with shuffled property labels.
In the ``\texttt{\#}Atoms'' baseline, the properties are predicted solely based on the number of atoms in the molecule.
Furthermore, we report the error of $\phi_{p}$ on $\sD_{b}$ as a lower bound baseline ``L-Bound''.

\textbf{Results.}\label{paragraph-results}
We generate molecules with textual descriptions targeted to each one of the six properties in QM9, which are detailed in the supplementary information.
As presented in Figure \ref{figure-MAE}, our TextSMOG has a lower MAE than other baselines on five out of the six properties, suggesting that the molecules generated by TextSMOG align more closely with the desired properties than other baselines.
The result underscores the proficiency of TextSMOG in exploiting textual data to guide the conditional \textit{de novo} generation of molecules.
Moreover, it highlights the superior congruence of the text-guided molecule generation via the diffusion model with the desired property, thus showing significant potential.
Furthermore, as indicated in Figure \ref{figure-NS}, our proposed TextSMOG exhibits commendable performance in terms of novelty and stability.
The text guidance we introduced has transformed the exploration of the model in the molecule generation space, generally enhancing the novelty of the generated molecules while maintaining their stability.

\begin{figure}
    \centering

    \includegraphics[]{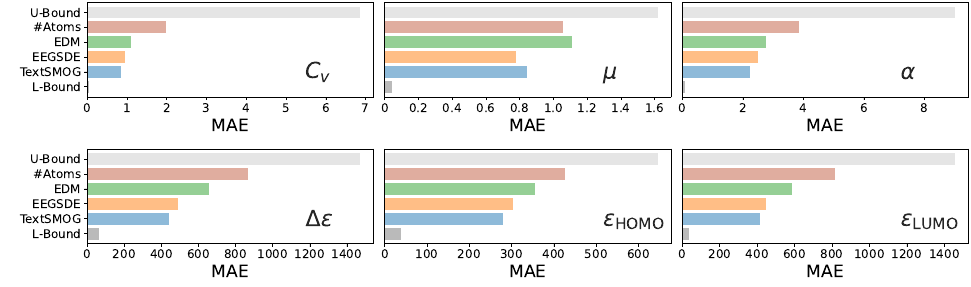}

    \caption{Comparison of MAE for the Generated Molecules Targeted to Desired Property. Statistics of baselines are from their original papers. The performance of EEGSDE varies depending on the scaling factor, and we report its best results. The numerical values are provided in the supplementary information.}
    \label{figure-MAE}
\end{figure}

\begin{figure}
    \centering
    \includegraphics[]{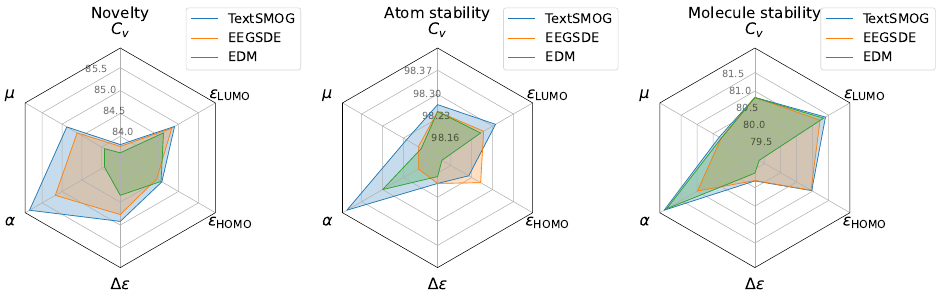}

    \caption{Comparison of Novelty (Novel, \%), Atom Stability (A. Stable,\%), and Molecule Stability (M. Stable,\%) on Generated Molecules Targeted to the Desired Property. Statistics of baselines are from EEGSDE. The performance of EEGSDE varies depending on the scaling factor, and we report its best results. }
    \label{figure-NS}
\end{figure}

\subsection{Experiment on Multiple Quantum Properties Conditioning}\label{subsection-exp_multi}
The capacity to generate molecules, guided by multiple conditions, is a crucial aspect of the molecule generation model.
When guided by textual descriptions, characterizing the condition with multiple desired properties is highly intuitive and flexible.
Following the same setup and metrics in the previous section, we evaluate our TextSMOG on the task of generating molecules with multiple desired quantum properties in QM9.
Then we compare TextSMOG with two baselines to showcase the effectiveness of our model in generating molecules conditioned on multiple quantum properties.

\begin{table}[t]
  \caption{Comparison of MAE on the Generated Molecules Targeted to the Multiple Desired Properties. Statistics of baselines are from EEGSDE. \best{Boldface} indicates the best performance. }
  \label{table-Multi-MAE}
  \begin{center}
          \begin{tabular}{@{}lcc@{}}
    \toprule
    Method      & MAE1$\downarrow$ & MAE2$\downarrow$ \\ \midrule
    Condition   & \multicolumn{2}{c}{$C_{v}\ \left(\frac{\textnormal{cal}}{\textnormal{mol}}\textnormal{K}\right)$ and $\mu\ (\textnormal{D})$} \\ \midrule
    EDM         &       1.079\var{0.007} &       1.156\var{0.011}  \\
    EEGSDE      &       0.981\var{0.008} &       0.912\var{0.006}  \\
    TextSMOG    & \best{0.645}\var{0.014} & \best{0.836}\var{0.017} \\ \midrule
    Condition   & \multicolumn{2}{c}{$\alpha\ (\textnormal{Bohr}^3)$ and $\mu\ (\textnormal{D})$} \\ \midrule
    EDM         &         2.76\var{0.01} &       1.158\var{0.002}  \\
    EEGSDE      &         2.61\var{0.01} &       0.855\var{0.007}  \\
    TextSMOG      & \best{2.27}\var{0.01} & \best{0.809}\var{0.010} \\ \midrule
    Condition   & \multicolumn{2}{c}{$\Delta\varepsilon\ (\textnormal{meV})$ and $\mu\ (\textnormal{D})$} \\ \midrule
    EDM         &       683\var{1} &       1.130\var{0.007}  \\
    EEGSDE      &       563\var{3} &       0.866\var{0.003}  \\
    TextSMOG    & \best{489}\var{4} & \best{0.843}\var{0.009} \\
    \bottomrule
\end{tabular}
  \end{center}
\end{table}

\begin{table}[t]
  \caption{Comparison of Novelty (Novel, \%), Atom Stability (A. Stable,\%), and Molecule Stability (M. Stable,\%) on the Generated Molecules Targeted to the Multiple Desired Properties. Statistics of baselines are from EEGSDE. \best{Boldface} indicates the best performance. }
  \label{table-Multi-NS}
  \begin{center}
          \begin{tabular}{@{}lccc@{}}
    \toprule
    Method      & Novel$\uparrow$ & A. Stable$\uparrow$ & M. Stable$\uparrow$  \\ \midrule
    Condition   & \multicolumn{3}{c}{$C_{v}\ \left(\frac{\textnormal{cal}}{\textnormal{mol}}\textnormal{K}\right)$ and $\mu\ (\textnormal{D})$} \\ \midrule
    EDM         &       85.31\var{0.43} & \best{98.00}\var{0.07} & \best{77.42}\var{0.80} \\
    EEGSDE      &       85.62\var{0.86} &       97.67\var{0.08} &       74.56\var{0.54} \\
    TextSMOG    & \best{85.79}\var{0.66} &       97.89\var{0.10} &       77.33\var{0.72} \\ \midrule
    Condition   & \multicolumn{3}{c}{$\alpha\ (\textnormal{Bohr}^3)$ and $\mu\ (\textnormal{D})$} \\ \midrule
    EDM         &       85.06\var{0.27} &       97.96\var{0.00} &       75.95\var{0.30} \\
    EEGSDE      &       85.56\var{0.56} &       97.61\var{0.04} &       72.72\var{0.27} \\
    TextSMOG    & \best{85.64}\var{0.64} & \best{98.01}\var{0.07} & \best{75.97}\var{0.44} \\ \midrule
    Condition   & \multicolumn{3}{c}{$\Delta\varepsilon\ (\textnormal{meV})$ and $\mu\ (\textnormal{D})$} \\ \midrule
    EDM         &       85.18\var{0.35} &       98.00\var{0.06} &       77.96\var{0.33} \\
    EEGSDE      &       85.36\var{0.03} &       97.99\var{0.06} &       77.77\var{0.26} \\
    TextSMOG    & \best{85.44}\var{0.41} & \best{98.06}\var{0.04} & \best{78.03}\var{0.29} \\ \bottomrule
\end{tabular}
  \end{center}
\end{table}

As shown in Table \ref{table-Multi-MAE}, our TextSMOG has a remarkably lower MAE than the other two baselines, thereby demonstrating the superiority of our model in generating molecules with multiple desired properties.
This also further substantiates that, without necessitating additional targeted interventions, textual conditions can be utilized in our model to guide molecule generation that conforms to multiple desired properties.

Additionally, as highlighted in Table \ref{table-Multi-NS}, our proposed TextSMOG maintains superior performance in terms of novelty and stability, when generating molecules targeted at multiple desired properties.
The results indicate that the flexible integration of multiple conditions through textual description does not compromise the stability of the generated molecules. Furthermore, this approach enhances novelty when compared to the baseline.

\subsection{Generation on General Textual Descriptions}
To further assess our model, we undertake additional training on a vast dataset of over 330K text-molecule pairs we gleaned from PubChem \citep{PubChem}. Then, we generate molecules based on general textual descriptions to observe the capacity of our model to generate from generalized textual conditions.

\begin{figure}[t]
  \centering
    \includegraphics[]{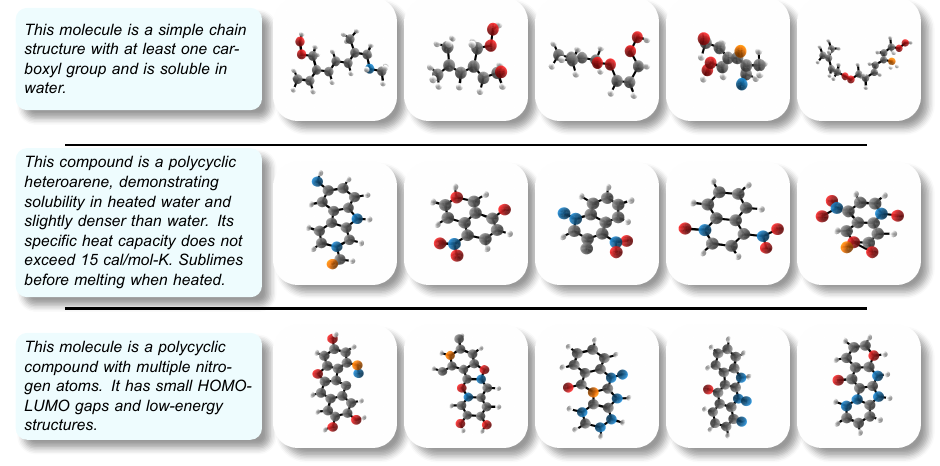}

  \caption{Generated molecules targeted to text description excerpts.}
  \label{figure-text}
\end{figure}

Visual observations, as depicted in Figure \ref{figure-text}, illuminate the impressive aptitude of our TextSMOG in aligning molecule structures with the desired property within the textual descriptions. For instance, when the textual description includes affirmatively mentioned terms such as "simple chain structure", "at least one carboxyl group", and "soluble in water", the generated molecules consistently exhibit chain structures with at least one carboxyl group, and characteristics indicative of water solubility.

Moreover, when the textual description includes "polycyclic heteroarene" and specifies the solubility and heat capacity of the molecule, TextSMOG generates a variety of polycyclic aromatic hydrocarbon molecules. The ubiquitously present amino and nitro groups attest to a certain degree of solubility of the molecules. Referring to structurally similar molecules, their expected specific heat capacity is also relatively low.

Lastly, when the text description explicitly demands multiple nitrogen atoms and a low energy gap, the molecules generated by TextSMOG not only possess the required polycyclic structure and multiple nitrogen atoms, but the rings on the same plane denote the low-energy structures of these molecules that are difficult to excite.

The remarkable alignment between the conditions and the generated molecule stands as a testament to the exceptional generative capabilities of TextSMOG. The result demonstrates that TextSMOG is equipped to deeply explore the chemical molecular space in a text-guided manner, thereby generating prospective molecules for subsequent applications. This capability could potentially expedite drug design and the discovery of materials.

The results highlight TextSMOG's versatility in generating a wide variety of molecular structures, from simple chain structures to complex polycyclic compounds, under the guidance of general text descriptions. This underscores the model's potential to perform well even when the conditions deviate significantly from the distribution of the training set.
		\section{DISCUSSION}

The translational impacts of TextSMOG are particularly significant for the field of drug discovery and materials science. By enabling the generation of molecular structures directly from textual descriptions, TextSMOG can streamline the early stages of drug design where rapid prototyping and iterative testing are crucial. This approach can facilitate the discovery of subject drug candidates by allowing researchers to quickly generate and evaluate provided molecules based on specific desired properties mentioned in literature or derived from expert knowledge.

Furthermore, TextSMOG can aid in the development of materials with tailored properties by generating molecules that meet specific criteria. This capability is valuable in industries such as polymers, nanomaterials, and catalysts, where precise molecular structures can significantly influence material performance.

In drug discovery, TextSMOG's ability to generate molecules that align with complex textual prompts can accelerate the identification of compounds with potential therapeutic effects. This is especially relevant for targeting diseases with well-characterized biochemical pathways, where detailed descriptions of molecular interactions and desired properties are available. By generating candidate molecules that meet these criteria, TextSMOG can help narrow down the pool of potential drugs, reducing the time and cost associated with experimental validation.

Additionally, TextSMOG's flexibility in handling diverse textual inputs can facilitate interdisciplinary research, where insights from different fields can be integrated into the molecule generation process. For instance, combining insights from biology, chemistry, and pharmacology can lead to more informed and effective drug design strategies.

Despite the complexity of translating textual prompts into accurate molecular structures, we have successfully integrated advanced language models with high-fidelity diffusion models in TextSMOG, a text-guided diffusion approach for 3D molecule generation. Our experiments on the QM9 and PubChem datasets demonstrate the superior performance of TextSMOG over leading baselines, affirming its efficacy in capturing desired properties from textual descriptions and generating corresponding valid molecules.

\section{LIMITATIONS OF STUDY}

The integration of textual information with the denoising process of a pre-trained equivariant diffusion model allows TextSMOG to generate valid and stable molecular conformations that closely align with diverse textual directives. This initial success paves the way for significant advancements in the exploration of chemical space and the development of compounds. Nevertheless, our findings are not without limitations.

Our work was constrained by the scarcity of high-quality data linking real-world 3D molecules to their corresponding textual descriptions. This limitation impacted our ability to fully train the model on a diverse set of text-3D molecule pairs, potentially affecting the accuracy of the generated molecules in generating molecules that accurately align with complex textual descriptions. Moreover, the relative slowness of the sampling process due to the iterative nature of the total diffusion steps can pose a challenge in scenarios requiring rapid molecule generation, such as high-throughput drug discovery or material design.

In addition to these limitations, the current design of TextSMOG necessitates that the properties to condition on must be known upfront during the training phase. This might not always be feasible in practical settings, where specific properties linked to a particular drug discovery target may only become available later on, and often with very limited sample data. The generalization of TextSMOG to more complex and real-world scenarios also needs further exploration.

Looking ahead, we are optimistic about the potential of text-guided 3D molecule generation to revolutionize drug discovery and related fields. Future work will focus on overcoming these challenges by expanding and enhancing the quality of datasets linking textual descriptions to molecular structures, improving the efficiency of the sampling process, and making TextSMOG more adaptable to real-world applications. Addressing these limitations will not only enhance the performance of TextSMOG but also contribute significantly to the advancement of text-guided molecule generation technology.

		\subsection{Acknowledgements}
		This research is supported by the National Natural Science Foundation of China (92270114).
    This research was also supported by the advanced computing resources provided by the Supercomputing Center of the USTC.

		\subsection{Author Contributions}
		Conceptualization, Yanchen Luo and Junfeng Fang; Methodology, Yanchen Luo and Sihang Li; Investigation,
		Zhiyuan Liu and Sihang Li; Writing - Original Draft, Yanchen Luo and Junfeng Fang;  Writing -
		Review \& Editing, Jiancan Wu, An zhang and Wenjie Du; Funding Acquisition, Xiang Wang; Resources,
		Jiancan Wu, An zhang and Wenjie Du; Supervision, Xiang Wang.

		\subsection{Competing Interests}
		The authors declare no competing interests.

    \clearpage
    \section{STAR Methods}

\subsection{Lead Contact}

Further information and requests for resources and reagents should be directed to and will be fulfilled by the lead contact, Yanchen Luo (luoyanchen@mail.ustc.edu.cn).

\subsection{Materials Availability}

This study did not generate new unique reagents.

\subsection{Data and Code Availability}

\begin{itemize}
    \item The datasets generated during this study are available at HuggingFace and are publicly available as of the date of publication. The url is listed in the key resources table.
    \item All original code has been deposited at GitHub and is publicly available as of the date of publication. The url is listed in the key resources table.
    \item Any additional information required to reanalyze the data reported in this paper is available from the lead contact upon request.
\end{itemize}

\subsection{Method Details}

In this section, we elaborate on the proposed text-guided small molecule generation approach via diffusion model (TextSMOG), as illustrated in Figure \ref{figure-framework}.
It integrates the textual information (\ie text guidance) into the conditional signal of diffusion models by employing the reference geometry that is described in the first subsection following.
Subsequently, we introduce an efficient learning approach that incorporates both the encoded conditional signal and pre-trained unconditional signal in the reverse process, to generate molecules that are not only structurally stable and chemically valid but also align well with the specified conditions, as presented in the second subsection.

\subsubsection{Notation and Background}
We begin with a background of diffusion-based 3D molecule generation, introducing the fundamental concepts of the diffusion model and delving into equivariant diffusion models.
See the comprehensive literature review on these topics in the Section Related works in Supplementary Information.
In accordance with prior studies \citep{EDM, EEGSDE, MDM}, we use the variable $\gG = (\vx, \vh)$ to represent the 3D molecular geometry.
Here $\vx = (x_{1}, \dots, x_{M}) \in \sR^{M \times 3}$ signifies the atom coordinates, while $\vh = (h_{1}, \dots, h_{M}) \in \sR^{M \times k}$ denotes the atom features.
These features encompass atom types and atom charges, characterizing the atomic properties within the molecular structure.

\paragraph{Diffusion Model}
The diffusion model \citep{Diffusion, DDPM} emerges as a leading generative model, having achieved great success in various domains \citep{ADM, LDMs, DreamBooth, SDE, SR3, ArchiSound}.
Typically, it is formulated as two Markov chains:
a forward process (\aka noising process) that gradually injects noise into the data, and a reverse process (\aka denoising process) that learns to recover the original data.
Such a reverse process endows the diffusion model with enhanced capabilities for effective data generation and recovery.

\textbf{Forward Process.}
Given the real 3D molecular geometry $\gG_0$, the forward process yields a sequence of intermediate variables $\gG_1, \cdots, \gG_T$ using the transition kernel $q(\gG_t|\gG_{t-1})$ in alignment with a variance schedule $\beta_1, \beta_2, \ldots, \beta_T \in (0, 1)$. Formally, it is expressed as:
\begin{equation}
    q(\gG_t | \gG_{t-1}) = \mathcal{N}(\gG_t| \sqrt{1 - \beta_t} \gG_{t-1}, \beta_t \mI_n),
\end{equation}
where $\mathcal{N}(\cdot|\cdot,\cdot)$ is a Gaussian distribution and $\mI_n$ is the identity matrix. This defines the joint distribution of $\gG_1, \cdots, \gG_T$ conditioned on $\gG_0$ using the chain rule of the Markov process:
\begin{equation}
    q(\gG_{1}, \cdots, \gG_{T} | \gG_{0}) = \prod_{t=1}^{T} q(\gG_{t} | \gG_{t-1}).
\end{equation}
Let $\alpha_t = 1 - \beta_t$ and $\bar{\alpha}_t := \prod_{s=1}^{t} \alpha_s$. The sampling of $\gG_t$ at time step $t$ is in a closed form:
\begin{equation}
    q(\gG_t | \gG_0) = \mathcal{N}(\gG_t| \sqrt{\bar{\alpha}_t} \gG_0, (1 - \bar{\alpha}_t) \mI_n).
\end{equation}
Accordingly, the forward process posteriors, when conditioned on $\gG_0$, are tractable as:
\begin{equation}
    q(\gG_{t-1} | \gG_t, \gG_0) = \mathcal{N}(\gG_{t-1}| \widetilde{\mu}(\gG_t, \gG_0), \widetilde{\beta}_t \mI_n),
\end{equation}
where
\begin{equation}
    \widetilde{\mu}(\gG_t, \gG_0) = \frac{\sqrt{\bar{\alpha}_{t-1}}\beta_t}{1 - \bar{\alpha}_t} \gG_0 + \frac{\sqrt{\alpha_t}(1 - \bar{\alpha}_t)}{1 - \bar{\alpha}_t} \gG_t, \quad
    \widetilde{\beta}_t = \frac{1 - \bar{\alpha}_{t-1}}{1 - \bar{\alpha}_t} \beta_t.
\end{equation}

\textbf{Reverse Process.}
To recover the original molecular geometry $\gG_0$, the diffusion model starts by generating a standard Gaussian noise $\gG_{T} \sim \mathcal{N}(\mO, \mI_n)$, then progressively eliminates noise through a reverse Markov chain.
This is characterized by a learnable transition kernel $p_\theta(\gG_{t-1} | \gG_{t})$ at each reverse step $t$, defined as:
\begin{equation}
    p_\theta(\gG_{t-1} | \gG_{t}) = \mathcal{N}(\gG_{t-1}| \mu_{\theta}(\gG_t, t), \Sigma_{\theta} (\gG_t, t)),
\end{equation}
where the variance $\Sigma_{\theta} (\gG_t, t) = \widetilde{\beta}_t \mI_{n}$ and the mean $\mu_{\theta}(\gG_t, t)$ is parameterized by deep neural networks with parameters $\theta$:
\begin{equation}
    \mu_{\theta}(\gG_t, t) = \widetilde{\mu}_{t}(\gG_t, \frac{1}{\sqrt{\bar{\alpha}_t}}(\gG_t - \sqrt{1 - \bar{\alpha}_t}\epsilon_{\theta}(\gG_t, t))) = \frac{1}{\sqrt{\alpha_t}} (\gG_t - \frac{1 - \alpha_t}{\sqrt{1 - \bar{\alpha}_t}} \epsilon_{\theta}(\gG_t, t)),
\end{equation}
where $\epsilon_{\theta}$ is a noise prediction function to approximate the noise $\epsilon$ from $\gG_{t}$.

With the reverse Markov chain, we can iteratively sample from the learnable transition kernel $p_\theta(\gG_{t-1} | \gG_{t})$ until $t=1$ to estimate the molecular geometry $\gG_0$.

\paragraph{Equivariant diffusion models}

The molecular geometry $\gG = (\vx, \vh)$ is inherently symmetric in 3D space --- that is, translating or rotating a molecule does not change its underlying structure or features.
Previous studies \citep{TFN, SE3-Transformers, LieConv} underscore the significance of leveraging these invariances in molecular representation learning for enhanced generalization.
However, the transformation of these higher-order representations usually requires computationally expensive approximations or coefficients \citep{EGNN, EDM}.
In contrast, equivariant diffusion models \citep{EF, EDM, GeoDiff} provide a more efficient approach to ensure both rotational and translational invariance.
The approach rests on the assumption that, with the model distribution $p(\gG) = p(\vx, \vh)$ remaining invariant to the Euclidean group E(3), identical molecules, despite being in different orientations, will correspond to the same distribution.
Based on this assumption, translational invariance is achieved by predicting only the deviations in coordinate with a zero center of mass, \ie $\sum_{i=1}^{M} x_{i} = 0$.
On the other hand, rotational invariance is accomplished by making the noise prediction network $\epsilon_{\theta}(\cdot)$ equivariant to orthogonal transformations \citep{EGNN, EDM}.
Specifically, given an orthogonal matrix $\mR$ representing a coordinate rotation or reflection, the conformation output $a^{\vx}$ from the network $\epsilon_{\theta}(\gG) = \epsilon_{\theta}(\vx, \vh) = (a^{\vx}, a^{\vh})$ is equivariant to $\mR$, if the following condition holds for all orthogonal matrices $\mR$:
\begin{equation}
    \epsilon_{\theta}(\mR \vx, \vh) = (\mR a^{\vx}, a^{\vh}).
\end{equation}

A model exhibiting rotational and translational equivariance means a neural network $p_{\theta}(\gG)$ can avoid learning orientations and translations of molecules from scratch \citep{EDM, EGNN}.
In this paper, we parameterize the noise prediction network $\epsilon_{\theta}$ using an E(n) equivariant graph neural network as introduced by \citep{EGNN}, which is a type of Graph Neural Network \citep{GNN} that satisfies the above equivariance constraint to E(3).

\subsubsection{Equivariant Diffusion Model for Molecule Generation}

Diffusion models, formulated as two Markov chains—a forward process that gradually injects noise into the data and a reverse process that learns to recover the original data—have been successfully applied to various domains, including molecule generation. This process is particularly effective in the context of molecule generation, where the forward process adds noise to the molecular geometry at each step until it is fully noise-corrupted. The reverse process then gradually denoises the initial geometry $\gG_{T}$ to generate the final molecular geometry $\gG_{0}$.

However, molecular geometries are inherently symmetric in 3D space—translations or rotations do not change their underlying structure or features. To take advantage of these invariances for improved generalization, we employ an equivariant diffusion model (EDM). The EDM ensures both rotational and translational invariance by predicting only the deviations in coordinate with a zero center of mass and making the noise prediction network $\epsilon_{\theta}(\cdot)$ equivariant to orthogonal transformations. This allows the model distribution $p(\gG)$ to remain invariant to the Euclidean group E(3), meaning identical molecules in different orientations correspond to the same distribution.

In this work, the integration of textual information into the conditional signal of the equivariant diffusion model is achieved by employing a reference geometry $\vc_{\textbf{P}}$ that is updated at each step based on the textual prompt $\textbf{P}$.

\subsubsection{Integrating Textual Prompts into 3D Molecular Reference Geometry}

To ensure high-fidelity 3D molecule generation, the reverse process of the diffusion model is typically guided by tailored conditional information representing desired properties like unique polarizability.
We represent this conditional information as $c$, which allows us to formulate the conditional reverse process as:
\begin{equation}
p_{\theta}(\gG_{t-1} | \gG_{t}, c) = \mathcal{N}(\gG_{t-1}| \mu_{\theta}(\gG_t, c, t), \widetilde{\beta}_t \mI_{n})
\end{equation}

Unlike previous approaches relying on limited value guidance (\ie property values), in this work, we aim to steer the reverse process with text guidance (\ie informative textual descriptions), which can convey a broader range of conditional requirements.
Intuitively, utilizing textual descriptions to specify conditional generation criteria not only provides greater expressivity but also better aligns the resulting 3D molecules with diverse and complex expectations.

Practically, we first introduce a textual prompt \textbf{P} describing desired 3D molecule properties. A multi-modal conversion module $\Gamma$, pre-trained on 300K text-molecule pairs from PubChem, is then employed. This module is comprised of a GIN molecular graph encoder \citep{GIN, GraphMVP} and a language encoder-decoder extended from BERT \citep{BERT, KVPLM}. It converts \textbf{P} into a reference geometry $\vc_{\textbf{P}}$, extracting specific information from the target conditions and refining the textual condition signal:
\begin{equation}
    \vc_{\textbf{P}} = \Gamma(\textbf{P}).
\end{equation}

Nevertheless, we should emphasize that valid and stable 3D molecules can hardly be obtained directly from $\vc_{\textbf{P}}$. The chemical fidelity in 3D molecular space may not be guaranteed. In what follows, we describe how to utilize $\vc_{\textbf{P}}$ for conditioning a pre-trained diffusion model to generate molecules that align with the desired properties, meanwhile alleviating the exhaustive training from scratch.

\subsubsection{Conditioning with the Reference of Text Guidance}
To leverage $\vc_{\textbf{P}}$ for text-guided conditional generation while preserving the validity and stability of the synthesized molecule, TextSMOG employs the iterative latent variable refinement (ILVR) \citep{ILVR} to condition a pre-trained unconditional diffusion model meanwhile maintaining inherent domain knowledge in the unconditional model.

With the pre-trained unconditional diffusion model EDM \citep{EDM}, we could perform a step-by-step reverse process.
Formally, at step $t$, we can sample an unconditional proposal molecular geometry:
\begin{equation}
    \widetilde{\gG}_{t - 1} \sim \widetilde{p}_{\widetilde{\theta}}(\widetilde{\gG}_{t - 1} | \gG_{t}).
\end{equation}
where $\widetilde{\theta}$ is the fixed parameters of the pre-trained unconditional diffusion model \citep{EDM}.
Then, to incorporate the condition signal $\vc_{\textbf{P}}$ in the reverse process, we introduce a linear operation $\varphi_{\theta}(\cdot)$.
Therefore the conditional denoising for one step at step $t$ can be formulated as:
\begin{equation}\label{equation-refine}
    \gG_{t - 1} = \varphi_{\theta}(\vc_{\textbf{P}}) + (\mathcal{I} - \varphi_{\theta}) (\widetilde{\gG}_{t - 1}),
\end{equation}
where $\mathcal{I}(\cdot)$ is the identity operation and $(\mathcal{I} - \varphi_{\theta})(\cdot)$ is the residual operation \wrt $\varphi_{\theta}(\cdot)$ \citep{james1971factorization}.
Accordingly, the condition signal $\vc_{\textbf{P}}$ is projected into the reverse denoising process by $\varphi_{\theta}(\cdot)$, thus $\gG_{t - 1}$ is obtained as the generated 3D molecular geometry conditioned on $\vc_{\textbf{P}}$.
Conceptually, the proposal geometry from unconditional generation $\widetilde{\gG}_{t - 1}$ tries to push the atoms into a chemically valid position, while the reference geometry $\vc_{\textbf{P}}$ pulls the atoms towards the structure targeted to the condition.

By matching latent variables following \Eqref{equation-refine}, we enable text-guided conditional generation with the unconditional diffusion model.
Accordingly, the one-step denoising distribution conditioned on textual guidance at each step $t$ can be reformulated as:
\begin{equation}
    \gG_{t - 1} \sim p_{\theta}(\gG_{t - 1} | \gG_{t}, \vc_{\textbf{P}}).
\end{equation}

\subsubsection{Training Objective}

To guarantee the quality of the generated molecules, the key lies in optimizing the variational lower bound (ELBO) of negative log-likelihood, which equals minimizing the Kullback-Leibler divergence between the joint distribution of the reverse Markov chain $p_{\theta}(\gG_0,\gG_1,\cdots,\gG_T)$ and the forward process $q(\gG_0,\gG_1,\cdots,\gG_T)$:
\begin{equation}
     \mathbb{E} \left[-\log p_{\theta} (\gG_0 | \vc_{\textbf{P}})\right] \le -\log\sum_{t \geq 1} \underbrace{\KL \left( q(\gG_{t-1} | \gG_t, \gG_0)||p_\theta(\gG_{t-1} | \gG_{t}, \vc_{\textbf{P}}) \right)}_{:=\mathcal{L}_{t-1}} + C,
\end{equation}
where $C$ is a constant independent of $\theta$.

Note that we set $\mathcal{L}_0 = - \log{p_{\theta}(\gG_0 | \gG_1 )}$ as a discrete decoder following \citep{DDPM}.
Further adopting the reparameterization from \citep{DDPM}, $\mathcal{L}_{t-1}$ can be simplified to:
\begin{equation}
    \mathcal{L}_{t-1} = \mathbb{E}_{\text{P}, \gG_0, \epsilon} \left[||\epsilon - \epsilon_\theta(\sqrt{\bar{\alpha}_t} \gG_0 + \sqrt{1-\bar{\alpha}_t}\epsilon, t, \vc_{\textbf{P}})||^2\right].
\end{equation}

\subsubsection{Evaluation metrics}\label{metrics}
\textbf{Mean absolute error (MAE).} \citep{MAE} is a measure of errors between paired observations.
Given the property classifier network $\phi_{p}$,  and the set of generated molecules $\sG$, the MAE is defined as:
\begin{equation}
    \textnormal{MAE} = \frac{1}{|\sG|} \sum_{\gG \in \sG} |\phi_{p}(\gG) - c_{\gG}|,
\end{equation}
where $\gG$ is the generated molecule, and of which $c_{\gG}$ is the desired property.

\textbf{Novelty.} \citep{GraphVAE} is the proportion of generated molecules that do not appear in the training set.
Specifically, let $\sG$ be the set of generated molecules, the novelty in our experiment is calculated as:
\begin{equation}
    \textnormal{Novelty} = \frac{|\sG \cap \sD_{b}|}{|\sG|}.
\end{equation}

\textbf{Atom stability.} \citep{EDM} is the proportion of the atoms in the generated molecules that have the right valency.
Specifically, the atom stability in our experiment is calculated as:
\begin{equation}
    \textnormal{Atom Stability} = \frac{\sum_{\gG \in \sG} |\sA_{\gG, \textnormal{stable}}|}{\sum_{\gG \in \sG} |\sA_{\gG}|},
\end{equation}
where $\sA_{\gG}$ is the set of atoms in the generated molecule $\gG$, and $\sA_{\gG, \textnormal{stable}}$ is the set of atoms in $\sA_{\gG}$ that have the right valency.

\textbf{Molecule stability.} \citep{EDM} is the proportion of the generated molecules where all atoms are stable.
Specifically, the molecule stability in our experiment is calculated as:
\begin{equation}
    \textnormal{Molecule Stability} = \frac{|\sG_{\textnormal{stable}}|}{|\sG|},
\end{equation}
where $\sG_{\textnormal{stable}}$ is the set of generated molecules where all atoms have the right valency.

\subsection{Quantification and Statistical Analysis}
\subsubsection{The Quantum Properties in QM9 Dataset}\label{dataset-QM9}
We consider 6 main quantum properties in QM9:
\begin{itemize}
    \item $C_{v}$: Heat capacity at 298.15K.
    \item $\mu$: Dipole moment.
    \item $\alpha$: Polarizability, which represents the tendency of a molecule to acquire an electric dipole moment when subjected to an external electric field.
    \item $\varepsilon_{\textnormal{HOMO}}$: Highest occupied molecular orbital energy.
    \item $\varepsilon_{\textnormal{LUMO}}$: Lowest unoccupied molecular orbital energy.
    \item $\Delta_{\varepsilon}$: The energy gap between HOMO and LUMO.
\end{itemize}\


    \section*{Supplementary Information}
    \section{Experiment Details}
\begin{table}[ht]
    \renewcommand{\thetable}{S1}
    \caption{Numerical Results of the Comparison of MAE in Figure 1. Statistics of baselines are from their original papers. The performance of EEGSDE varies depending on the scaling factor, and we report its best results. \best{Boldface} indicates the best performance.}
    \label{table-MAE}
    \begin{center}
            \begin{tabular}{@{}lcccccc@{}}
    \toprule
    \multirow{2}{*}{Method} & \multicolumn{6}{c}{MAE$\downarrow$}                                                                                                                                      \\ \cmidrule(l){2-7}
                            & \multicolumn{1}{c}{$C_{v}\ \left(\frac{\textnormal{cal}}{\textnormal{mol}}\textnormal{K}\right)$} & \multicolumn{1}{c}{$\mu\ (\textnormal{D})$} & \multicolumn{1}{c}{$\alpha\ (\textnormal{Bohr}\phantom{0}3)$} & \multicolumn{1}{c}{$\Delta\varepsilon\ (\textnormal{meV})$} & \multicolumn{1}{c}{$\varepsilon_{\textnormal{HOMO}}\ (\textnormal{meV})$} & \multicolumn{1}{c}{$\varepsilon_{\textnormal{LUMO}}\ (\textnormal{meV})$} \\ \midrule
    U-Bound                 &       6.857\var{0.0020} &       1.615\var{0.0004} &      9.00\var{0.03} &       1470\var{5} &       645\var{41} &       1457\var{5} \\
    \texttt{\#}Atoms        &       1.971\var{0.0000} &       1.053\var{0.0000} &      3.86\var{0.00} &       \phantom{0}866\var{0} &       426\var{0\phantom{0}} &       \phantom{0}813\var{0} \\
    EDM                     &       1.065\var{0.0010} &       1.123\var{0.0013} &      2.76\var{0.04} &       \phantom{0}655\var{8} &       356\var{5\phantom{0}} &       \phantom{0}584\var{7} \\
    EEGSDE                  &       0.941\var{0.0005} & \best{0.777}\var{0.0007} &      2.50\var{0.02} &       \phantom{0}487\var{3} &       302\var{2\phantom{0}} &       \phantom{0}447\var{6} \\
    TextSMOG                & \best{0.849}\var{0.0007} &       0.848\var{0.0010} &\best{2.24}\var{0.03} & \best{\phantom{0}443}\var{6} & \best{279}\var{4\phantom{0}} & \best{\phantom{0}412}\var{8} \\
    L-Bound                 &       0.040\var{0.0000} &       0.043\var{0.0000} &      0.10\var{0.00} &       \phantom{00}64\var{0} &       \phantom{0}39\var{0\phantom{0}} &       \phantom{00}36\var{0} \\ \bottomrule
\end{tabular}


    \end{center}
\end{table}

\begin{table}[h]
    \renewcommand{\thetable}{S2}
    \caption{As a supplement to Section Results, the comparison over 10000 generated molecules of models on unconditional generation with standard deviations across 3 runs on QM9. NLL: Negative log-likelihood.}
    \label{table-unconditional}
    \begin{center}
            \begin{tabular}{cccccc}
    \toprule
             & NLL    & Stability (\%) & Validity (\%) & Uniqueness (\%) & Novelty (\%) \\
    \midrule
    ENF      & -59.7  & 24.6 & 41.0 & 40.1 & 39.5 \\
    G-Schnet & -      & 85.6 & 85.9 & 80.9 & 57.6 \\
    EDM      & -110.7 & 91.1 & 91.9 & 90.7 & 89.9 \\
    MDM      & -      & 91.9 & 98.6 & 94.6 & 90.0 \\
    Ours     & -105   & 93.1 & 99.1 & 95.9 & 89.4 \\
    \bottomrule
\end{tabular}
    \end{center}
\end{table}

\begin{figure}
    \renewcommand{\thefigure}{S1}
    \centering
    \includegraphics[]{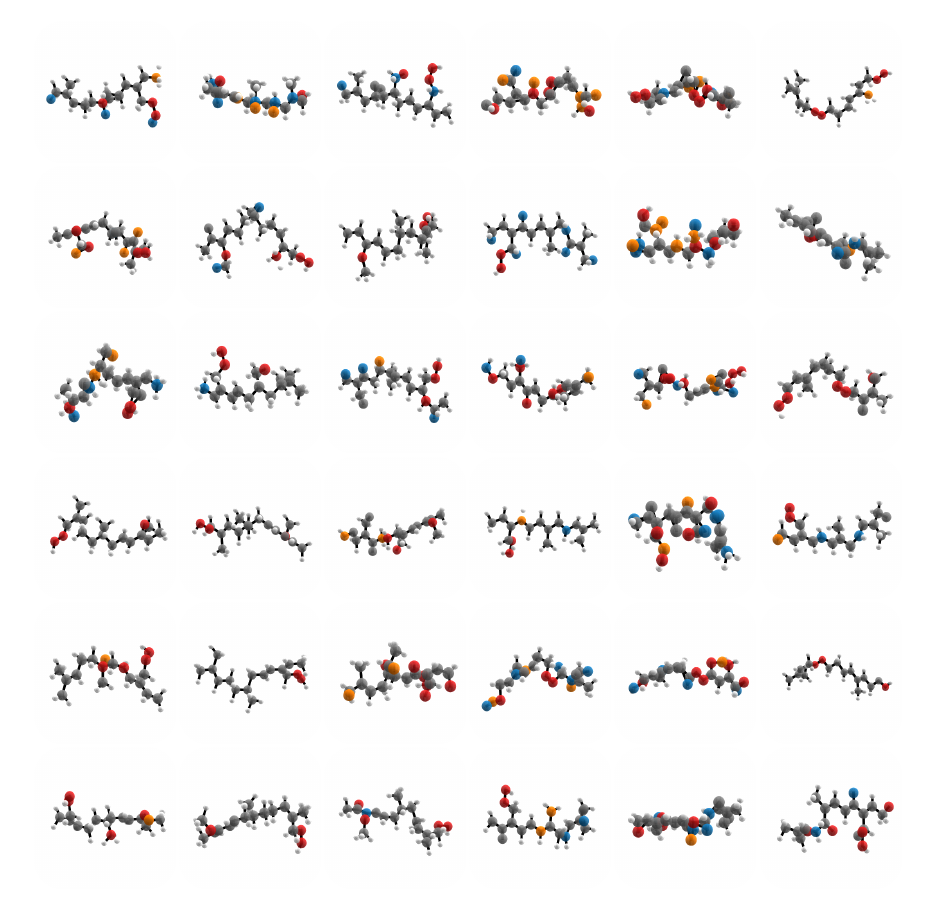}
    \caption{As a supplement to Figure 4, random examples of molecules generated for the text description, random examples of generated molecules targeted to text description "This molecule is a simple chain structure with at least one carboxyl group and is soluble in water."}
    \label{figure-examples-1}
\end{figure}

\begin{figure}
    \renewcommand{\thefigure}{S2}
    \centering
    \includegraphics[]{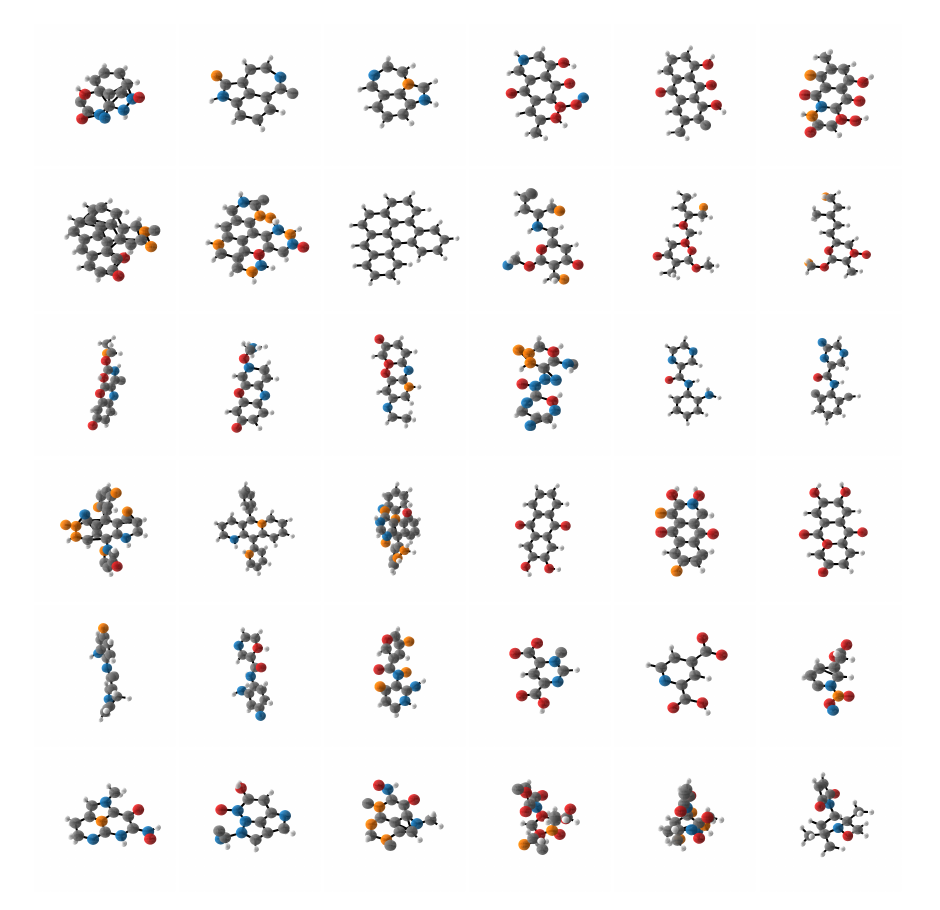}
    \caption{As a supplement to Figure 4, random examples of generated molecules targeted to text description "This molecule is a polycyclic compound with multiple nitrogen atoms. It has small HOMO-LUMO gaps and low-energy structures."}
    \label{figure-examples-2}
\end{figure}

    \section{Data S1}\label{sec:related-work}
\textbf{Diffusion models} are initially proposed by \citep{Diffusion}.
The basic idea is to corrupt data with diffusion noise and learn a neural diffusion model to reconstruct data from noise.
Recently, they have been theoretically enhanced by establishing connections to score matching and stochastic differential equations (SDE) \citep{DDPM, SDE}.
Such theoretical enhancements have facilitated the successful application of diffusion models across diverse domains, including image and waveform generation \citep{ADM, LDMs, WaveGrad, DiffWave}, and have recently gained attention in the molecular sciences field \citep{EDM, MDM, GeoDiff}.

\textbf{Molecule generation} is to explore the molecular space and generate subject molecules.
Prior efforts \citep{SMILES, cRNN, JT-VAE, E-GCL, SimSGT, MolCA} often generate simplified representations of molecules, such as 1D SMILES strings and 2D molecule graphs.
Some studies \citep{torsional_diffusion} have also tried to generate torsion angles in a given 2D molecular graph for the conformation generation task.
More recently, several works \citep{3DMolNet, G-SchNet, E-NFs, EDMnets, EDM} attempt to model molecules as 3D objects via deep generative models.
Diverse model architectures are proposed, including, but not limited to, variational autoencoders \citep{GVA, SD-VAE, JT-VAE, GraphVAE, CGVAE}, normalizing flows \citep{GraphNVP, MoFlow, GraphDF}, generative adversarial networks \citep{dcGAN, DEFactor}, autoregressive models \citep{GraphAF, MolecularRNN, Keeping_it_Simple}.
In the most recent developments, diffusion models have gained prominence in molecule generation \citep{EDM, EEGSDE, MDM, GeoDiff, Bridge}, marking a novel direction in the field.

Generally, these methods can be categorized into unconditional and conditional molecule generation.
Unconditional molecule generation \citep{EDM, MDM} generates molecules without any external constraints, representing the naive form of molecule generation.

\textbf{Conditional molecule generation}, however, which conduct valid molecules that exhibit desired properties \citep{SSVAE, cRNN, CMGN}, is a pivotal approach of inverse molecular design \citep{Inverse_molecular_design}.
Towards this end, many prior works \citep{EDM, G-SchNet, cG-SchNet} adopt the idea of conditional diffusion, having centered on learning a molecule distribution conditioned on certain properties from existing data.
By sampling from this distribution with conditions aligning with desired properties, subject molecules can be generated.
Here we scrutinize the widely-used condition types.
Many previous attempts \citep{EDM, EEGSDE, MDM} mostly employ a specific property value (\eg polarizability, dipole moment, and molecular orbital energy) as the condition in diffusion, ensuring the generated molecules adhere to the particular chemical or quantum attributes.
These efforts set value-based conditions to ensure the molecules conform to certain chemical or quantum characteristics.
Some studies \citep{cG-SchNet, cRNN} stipulate specific structural conditions as molecular fingerprints.
However, solely specifying a target property often falls short of addressing the comprehensive demands of inverse molecular design \citep{Honório_Moda_Andricopulo_2013, cG-SchNet, MGCVAE}.
To overcome this limitation, some studies \citep{cG-SchNet, EEGSDE, CMGN} have combined that combine multiple properties as conditions.
Such strategies can cater to multiple targets in inverse molecular design, such as generating molecules with low-energy structures and small HOMO-LUMO gaps.
In contrast to these value-based conditional generative models confined to a single or a handful of properties, our work further proposes a text-guided method, a flexible and generalized way to control the generation process of molecules.


    \bibliography{0main}

\begin{thebibliography}{83}
\expandafter\ifx\csname natexlab\endcsname\relax\def\natexlab#1{#1}\fi

\bibitem[{Alcalde et~al.(2006)Alcalde, Ferrer, Plou, and Ballesteros}]{Alcalde_Ferrer_Plou_2007}
Miguel Alcalde, Manuel Ferrer, Francisco~J. Plou, and Antonio Ballesteros. 2006.
\newblock \href {https://doi.org/https://doi.org/10.1016/j.tibtech.2006.04.002} {Environmental biocatalysis: from remediation with enzymes to novel green processes}.
\newblock \emph{Trends in Biotechnology}, 24(6):281--287.

\bibitem[{Anand et~al.(2022)Anand, Eguchi, Mathews, Perez, Derry, Altman, and Huang}]{anand2022protein}
Namrata Anand, Raphael Eguchi, Irimpan~I Mathews, Carla~P Perez, Alexander Derry, Russ~B Altman, and Po-Ssu Huang. 2022.
\newblock \href {https://doi.org/10.1038/s41467-022-28313-9} {Protein sequence design with a learned potential}.
\newblock \emph{Nature communications}, 13(1):746.

\bibitem[{Assouel et~al.(2018)Assouel, Ahmed, Segler, Saffari, and Bengio}]{DEFactor}
Rim Assouel, Mohamed Ahmed, Marwin H.~S. Segler, Amir Saffari, and Yoshua Bengio. 2018.
\newblock \href {https://doi.org/10.48550/arXiv.1811.09766} {Defactor: Differentiable edge factorization-based probabilistic graph generation}.
\newblock \emph{CoRR}, abs/1811.09766.

\bibitem[{Bao et~al.(2023)Bao, Zhao, Hao, Li, Li, and Zhu}]{EEGSDE}
Fan Bao, Min Zhao, Zhongkai Hao, Peiyao Li, Chongxuan Li, and Jun Zhu. 2023.
\newblock \href {https://doi.org/10.48550/arXiv.2209.15408} {Equivariant energy-guided {SDE} for inverse molecular design}.
\newblock In \emph{{ICLR}}. OpenReview.net.

\bibitem[{Barakat et~al.(2014)Barakat, Houghton, Tyrrell, and Tuszynski}]{Khaled_H_Barakat_2014}
Khaled~H. Barakat, Michael Houghton, D.~Lorne Tyrrell, and Jack~A. Tuszynski. 2014.
\newblock \href {https://doi.org/10.4018/ijcmam.2014010104} {Rational drug design: One target, many paths to it}.
\newblock \emph{International Journal of Computational Models and Algorithms in Medicine (IJCMAM)}, 4(1):59--85.

\bibitem[{Beltagy et~al.(2019)Beltagy, Lo, and Cohan}]{SciBERT}
Iz~Beltagy, Kyle Lo, and Arman Cohan. 2019.
\newblock \href {https://doi.org/10.48550/arXiv.1903.10676} {Scibert: {A} pretrained language model for scientific text}.
\newblock In \emph{{EMNLP/IJCNLP} {(1)}}, pages 3613--3618. Association for Computational Linguistics.

\bibitem[{Bian et~al.(2019)Bian, Wang, Jun, and Xie}]{dcGAN}
Yuemin Bian, Junmei Wang, Jaden~Jungho Jun, and Xiang-Qun Xie. 2019.
\newblock \href {https://doi.org/10.1021/acs.molpharmaceut.9b00500} {Deep convolutional generative adversarial network (dcgan) models for screening and design of small molecules targeting cannabinoid receptors}.
\newblock \emph{Molecular Pharmaceutics}, 16(11):4451--4460.

\bibitem[{Brown et~al.(2020)Brown, Mann, Ryder, Subbiah, Kaplan, Dhariwal, Neelakantan, Shyam, Sastry, Askell, Agarwal, Herbert{-}Voss, Krueger, Henighan, Child, Ramesh, Ziegler, Wu, Winter, Hesse, Chen, Sigler, Litwin, Gray, Chess, Clark, Berner, McCandlish, Radford, Sutskever, and Amodei}]{GPT-3}
Tom~B. Brown, Benjamin Mann, Nick Ryder, Melanie Subbiah, Jared Kaplan, Prafulla Dhariwal, Arvind Neelakantan, Pranav Shyam, Girish Sastry, Amanda Askell, Sandhini Agarwal, Ariel Herbert{-}Voss, Gretchen Krueger, Tom Henighan, Rewon Child, Aditya Ramesh, Daniel~M. Ziegler, Jeffrey Wu, Clemens Winter, Christopher Hesse, Mark Chen, Eric Sigler, Mateusz Litwin, Scott Gray, Benjamin Chess, Jack Clark, Christopher Berner, Sam McCandlish, Alec Radford, Ilya Sutskever, and Dario Amodei. 2020.
\newblock \href {https://doi.org/10.48550/arXiv.2005.14165} {Language models are few-shot learners}.
\newblock In \emph{NeurIPS}.

\bibitem[{Chen et~al.(2021)Chen, Zhang, Zen, Weiss, Norouzi, and Chan}]{WaveGrad}
Nanxin Chen, Yu~Zhang, Heiga Zen, Ron~J. Weiss, Mohammad Norouzi, and William Chan. 2021.
\newblock \href {https://doi.org/10.48550/arXiv.2009.00713} {Wavegrad: Estimating gradients for waveform generation}.
\newblock In \emph{{ICLR}}. OpenReview.net.

\bibitem[{Choi et~al.(2021)Choi, Kim, Jeong, Gwon, and Yoon}]{ILVR}
Jooyoung Choi, Sungwon Kim, Yonghyun Jeong, Youngjune Gwon, and Sungroh Yoon. 2021.
\newblock \href {https://doi.org/10.48550/arXiv.2108.02938} {{ILVR:} conditioning method for denoising diffusion probabilistic models}.
\newblock In \emph{{ICCV}}, pages 14347--14356. {IEEE}.

\bibitem[{Dai et~al.(2018)Dai, Tian, Dai, Skiena, and Song}]{SD-VAE}
Hanjun Dai, Yingtao Tian, Bo~Dai, Steven Skiena, and Le~Song. 2018.
\newblock \href {https://doi.org/10.48550/arXiv.1802.08786} {Syntax-directed variational autoencoder for structured data}.
\newblock In \emph{{ICLR} (Poster)}. OpenReview.net.

\bibitem[{Degtyarenko et~al.(2008)Degtyarenko, de~Matos, Ennis, Hastings, Zbinden, McNaught, Alc{\'{a}}ntara, Darsow, Guedj, and Ashburner}]{ChEBI}
Kirill Degtyarenko, Paula de~Matos, Marcus Ennis, Janna Hastings, Martin Zbinden, Alan McNaught, Rafael Alc{\'{a}}ntara, Michael Darsow, Micka{\"{e}}l Guedj, and Michael Ashburner. 2008.
\newblock \href {https://doi.org/10.1093/NAR/GKM791} {Chebi: a database and ontology for chemical entities of biological interest}.
\newblock \emph{Nucleic Acids Res.}, 36(Database-Issue):344--350.

\bibitem[{Devlin et~al.(2019)Devlin, Chang, Lee, and Toutanova}]{BERT}
Jacob Devlin, Ming{-}Wei Chang, Kenton Lee, and Kristina Toutanova. 2019.
\newblock \href {https://doi.org/10.48550/arXiv.1810.04805} {{BERT:} pre-training of deep bidirectional transformers for language understanding}.
\newblock In \emph{{NAACL-HLT} {(1)}}, pages 4171--4186. Association for Computational Linguistics.

\bibitem[{Dhariwal and Nichol(2021)}]{ADM}
Prafulla Dhariwal and Alexander~Quinn Nichol. 2021.
\newblock \href {https://doi.org/10.48550/arXiv.2105.05233} {Diffusion models beat gans on image synthesis}.
\newblock In \emph{NeurIPS}, pages 8780--8794.

\bibitem[{Edwards et~al.(2022)Edwards, Lai, Ros, Honke, Cho, and Ji}]{MolT5}
Carl Edwards, Tuan~Manh Lai, Kevin Ros, Garrett Honke, Kyunghyun Cho, and Heng Ji. 2022.
\newblock \href {https://doi.org/10.48550/arXiv.2204.11817} {Translation between molecules and natural language}.
\newblock In \emph{{EMNLP}}, pages 375--413. Association for Computational Linguistics.

\bibitem[{Edwards et~al.(2021)Edwards, Zhai, and Ji}]{Text2Mol}
Carl Edwards, ChengXiang Zhai, and Heng Ji. 2021.
\newblock \href {https://doi.org/10.18653/v1/2021.emnlp-main.47} {Text2mol: Cross-modal molecule retrieval with natural language queries}.
\newblock In \emph{{EMNLP} {(1)}}, pages 595--607. Association for Computational Linguistics.

\bibitem[{Fang et~al.(2024)Fang, Zhang, Wu, Yang, Liu, Li, Wang, Du, and Wang}]{MolTC}
Junfeng Fang, Shuai Zhang, Chang Wu, Zhengyi Yang, Zhiyuan Liu, Sihang Li, Kun Wang, Wenjie Du, and Xiang Wang. 2024.
\newblock Moltc: Towards molecular relational modeling in language models.
\newblock In \emph{{ACL} (Findings)}, pages 1943--1958. Association for Computational Linguistics.

\bibitem[{Finzi et~al.(2020)Finzi, Stanton, Izmailov, and Wilson}]{LieConv}
Marc Finzi, Samuel Stanton, Pavel Izmailov, and Andrew~Gordon Wilson. 2020.
\newblock \href {https://doi.org/10.48550/arXiv.2002.12880} {Generalizing convolutional neural networks for equivariance to lie groups on arbitrary continuous data}.
\newblock In \emph{{ICML}}, volume 119 of \emph{Proceedings of Machine Learning Research}, pages 3165--3176. {PMLR}.

\bibitem[{Flam{-}Shepherd et~al.(2022)Flam{-}Shepherd, Zhu, and Aspuru{-}Guzik}]{Keeping_it_Simple}
Daniel Flam{-}Shepherd, Kevin Zhu, and Al{\'{a}}n Aspuru{-}Guzik. 2022.
\newblock \href {https://doi.org/10.1038/s41467-022-30839-x} {Language models can learn complex molecular distributions}.
\newblock \emph{Nature Communications}, 13:3293.

\bibitem[{Fuchs et~al.(2020)Fuchs, Worrall, Fischer, and Welling}]{SE3-Transformers}
Fabian Fuchs, Daniel~E. Worrall, Volker Fischer, and Max Welling. 2020.
\newblock \href {https://doi.org/10.48550/arXiv.2006.10503} {Se3-transformers: 3d roto-translation equivariant attention networks}.
\newblock In \emph{NeurIPS}.

\bibitem[{Gaudelet et~al.(2021)Gaudelet, Day, Jamasb, Soman, Regep, Liu, Hayter, Vickers, Roberts, Tang, Roblin, Blundell, Bronstein, and Taylor{-}King}]{GML}
Thomas Gaudelet, Ben Day, Arian~R. Jamasb, Jyothish Soman, Cristian Regep, Gertrude Liu, Jeremy B.~R. Hayter, Richard Vickers, Charles Roberts, Jian Tang, David Roblin, Tom~L. Blundell, Michael~M. Bronstein, and Jake~P. Taylor{-}King. 2021.
\newblock \href {https://doi.org/10.1093/bib/bbab159} {Utilizing graph machine learning within drug discovery and development}.
\newblock \emph{Briefings in Bioinformatics}, 22(6):bbab159.

\bibitem[{Gebauer et~al.(2022)Gebauer, Gastegger, Hessmann, M{\"{u}}ller, and Sch{\"{u}}tt}]{cG-SchNet}
Niklas W.~A. Gebauer, Michael Gastegger, Stefaan S.~P. Hessmann, Klaus{-}Robert M{\"{u}}ller, and Kristof~T. Sch{\"{u}}tt. 2022.
\newblock \href {https://doi.org/10.1038/s41467-022-28526-y} {Inverse design of 3d molecular structures with conditional generative neural networks}.
\newblock \emph{Nature Communications}, 13:973.

\bibitem[{Gebauer et~al.(2019)Gebauer, Gastegger, and Sch{\"{u}}tt}]{G-SchNet}
Niklas W.~A. Gebauer, Michael Gastegger, and Kristof Sch{\"{u}}tt. 2019.
\newblock \href {https://doi.org/10.48550/arXiv.1906.00957} {Symmetry-adapted generation of 3d point sets for the targeted discovery of molecules}.
\newblock In \emph{NeurIPS}, pages 7564--7576.

\bibitem[{Hajduk and Greer(2007)}]{Hajduk_Greer_2007}
Philip~J. Hajduk and Jonathan Greer. 2007.
\newblock \href {https://doi.org/10.1038/nrd2220} {A decade of fragment-based drug design: strategic advances and lessons learned.}
\newblock \emph{Nature Reviews Drug Discovery}, 6:211–219.

\bibitem[{Hamilton et~al.(2017)Hamilton, Ying, and Leskovec}]{GNN}
William~L. Hamilton, Zhitao Ying, and Jure Leskovec. 2017.
\newblock \href {https://doi.org/10.48550/arXiv.1706.02216} {Inductive representation learning on large graphs}.
\newblock In \emph{{NIPS}}, pages 1024--1034.

\bibitem[{Ho et~al.(2020)Ho, Jain, and Abbeel}]{DDPM}
Jonathan Ho, Ajay Jain, and Pieter Abbeel. 2020.
\newblock \href {https://doi.org/10.48550/arXiv.2006.11239} {Denoising diffusion probabilistic models}.
\newblock In \emph{NeurIPS}.

\bibitem[{Hoffmann and No{\'{e}}(2019)}]{EDMnets}
Moritz Hoffmann and Frank No{\'{e}}. 2019.
\newblock \href {https://doi.org/10.48550/arXiv.1910.03131} {Generating valid euclidean distance matrices}.
\newblock \emph{CoRR}, abs/1910.03131.

\bibitem[{Honório et~al.(2013)Honório, Moda, and Andricopulo}]{Honório_Moda_Andricopulo_2013}
KáthiaMaria Honório, TiagoL. Moda, and AdrianoD. Andricopulo. 2013.
\newblock \href {https://doi.org/10.2174/1573406411309020002} {Pharmacokinetic properties and in silico adme modeling in drug discovery}.
\newblock \emph{Medicinal Chemistry,Medicinal Chemistry}, 9(2):163--176.

\bibitem[{Hoogeboom et~al.(2022)Hoogeboom, Satorras, Vignac, and Welling}]{EDM}
Emiel Hoogeboom, Victor~Garcia Satorras, Cl{\'{e}}ment Vignac, and Max Welling. 2022.
\newblock \href {https://doi.org/10.48550/arXiv.2203.17003} {Equivariant diffusion for molecule generation in 3d}.
\newblock In \emph{{ICML}}, volume 162 of \emph{Proceedings of Machine Learning Research}, pages 8867--8887. {PMLR}.

\bibitem[{Huang et~al.(2023)Huang, Zhang, Xu, and Wong}]{MDM}
Lei Huang, Hengtong Zhang, Tingyang Xu, and Ka{-}Chun Wong. 2023.
\newblock \href {https://doi.org/10.48550/arXiv.2209.05710} {{MDM:} molecular diffusion model for 3d molecule generation}.
\newblock In \emph{{AAAI}}, pages 5105--5112. {AAAI} Press.

\bibitem[{James and Wilkinson(1971)}]{james1971factorization}
AT~James and GN~Wilkinson. 1971.
\newblock \href {https://doi.org/10.2307/2334516} {Factorization of the residual operator and canonical decomposition of nonorthogonal factors in the analysis of variance}.
\newblock \emph{Biometrika}, 58(2):279--294.

\bibitem[{Jin et~al.(2018)Jin, Barzilay, and Jaakkola}]{JT-VAE}
Wengong Jin, Regina Barzilay, and Tommi~S. Jaakkola. 2018.
\newblock \href {https://doi.org/10.48550/arXiv.1802.04364} {Junction tree variational autoencoder for molecular graph generation}.
\newblock In \emph{{ICML}}, volume~80 of \emph{Proceedings of Machine Learning Research}, pages 2328--2337. {PMLR}.

\bibitem[{Jing et~al.(2022)Jing, Corso, Chang, Barzilay, and Jaakkola}]{torsional_diffusion}
Bowen Jing, Gabriele Corso, Jeffrey Chang, Regina Barzilay, and Tommi~S. Jaakkola. 2022.
\newblock \href {https://doi.org/10.48550/arXiv.2206.01729} {Torsional diffusion for molecular conformer generation}.
\newblock In \emph{NeurIPS}.

\bibitem[{Kang and Cho(2019)}]{SSVAE}
Seokho Kang and Kyunghyun Cho. 2019.
\newblock \href {https://doi.org/10.1021/acs.jcim.8b00263} {Conditional molecular design with deep generative models}.
\newblock \emph{Journal of Chemical Information and Modeling}, 59(1):43--52.

\bibitem[{Kim et~al.(2021)Kim, Chen, Cheng, Gindulyte, He, He, Li, Shoemaker, Thiessen, Yu, Zaslavsky, Zhang, and Bolton}]{PubChem}
Sunghwan Kim, Jie Chen, Tiejun Cheng, Asta Gindulyte, Jia He, Siqian He, Qingliang Li, Benjamin~A. Shoemaker, Paul~A. Thiessen, Bo~Yu, Leonid Zaslavsky, Jian Zhang, and Evan Bolton. 2021.
\newblock \href {https://doi.org/10.1093/NAR/GKAA971} {Pubchem in 2021: new data content and improved web interfaces}.
\newblock \emph{Nucleic Acids Res.}, 49(Database-Issue):D1388--D1395.

\bibitem[{K{\"{o}}hler et~al.(2020)K{\"{o}}hler, Klein, and No{\'{e}}}]{EF}
Jonas K{\"{o}}hler, Leon Klein, and Frank No{\'{e}}. 2020.
\newblock \href {https://doi.org/10.48550/arXiv.2006.02425} {Equivariant flows: Exact likelihood generative learning for symmetric densities}.
\newblock In \emph{{ICML}}, volume 119 of \emph{Proceedings of Machine Learning Research}, pages 5361--5370. {PMLR}.

\bibitem[{Kong et~al.(2021)Kong, Ping, Huang, Zhao, and Catanzaro}]{DiffWave}
Zhifeng Kong, Wei Ping, Jiaji Huang, Kexin Zhao, and Bryan Catanzaro. 2021.
\newblock \href {https://doi.org/10.48550/arXiv.2009.09761} {Diffwave: {A} versatile diffusion model for audio synthesis}.
\newblock In \emph{{ICLR}}. OpenReview.net.

\bibitem[{Kotsias et~al.(2020)Kotsias, Ar{\'{u}}s{-}Pous, Chen, Engkvist, Tyrchan, and Bjerrum}]{cRNN}
Panagiotis{-}Christos Kotsias, Josep Ar{\'{u}}s{-}Pous, Hongming Chen, Ola Engkvist, Christian Tyrchan, and Esben~Jannik Bjerrum. 2020.
\newblock \href {https://doi.org/10.1038/s42256-020-0174-5} {Direct steering of de novo molecular generation with descriptor conditional recurrent neural networks}.
\newblock \emph{Nature Machine Intelligence}, 2(5):254--265.

\bibitem[{Kusner et~al.(2017)Kusner, Paige, and Hern{\'{a}}ndez{-}Lobato}]{GVA}
Matt~J. Kusner, Brooks Paige, and Jos{\'{e}}~Miguel Hern{\'{a}}ndez{-}Lobato. 2017.
\newblock \href {https://doi.org/10.48550/arXiv.1703.01925} {Grammar variational autoencoder}.
\newblock In \emph{{ICML}}, volume~70 of \emph{Proceedings of Machine Learning Research}, pages 1945--1954. {PMLR}.

\bibitem[{Lee and Min(2022)}]{MGCVAE}
Myeonghun Lee and Kyoungmin Min. 2022.
\newblock \href {https://doi.org/10.1021/acs.jcim.2c00487} {{MGCVAE:} multi-objective inverse design via molecular graph conditional variational autoencoder}.
\newblock \emph{Journal of Chemical Information and Modeling}, 62(12):2943--2950.

\bibitem[{Li et~al.(2024)Li, Liu, Luo, Wang, He, Kawaguchi, Chua, and Tian}]{3DMoLM}
Sihang Li, Zhiyuan Liu, Yanchen Luo, Xiang Wang, Xiangnan He, Kenji Kawaguchi, Tat{-}Seng Chua, and Qi~Tian. 2024.
\newblock Towards 3d molecule-text interpretation in language models.
\newblock In \emph{{ICLR}}. OpenReview.net.

\bibitem[{Liu et~al.(2018)Liu, Allamanis, Brockschmidt, and Gaunt}]{CGVAE}
Qi~Liu, Miltiadis Allamanis, Marc Brockschmidt, and Alexander~L. Gaunt. 2018.
\newblock \href {https://doi.org/10.48550/arXiv.1805.09076} {Constrained graph variational autoencoders for molecule design}.
\newblock In \emph{NeurIPS}, pages 7806--7815.

\bibitem[{Liu et~al.(2022)Liu, Wang, Liu, Lasenby, Guo, and Tang}]{GraphMVP}
Shengchao Liu, Hanchen Wang, Weiyang Liu, Joan Lasenby, Hongyu Guo, and Jian Tang. 2022.
\newblock \href {https://doi.org/10.48550/arXiv.2110.07728} {Pre-training molecular graph representation with 3d geometry}.
\newblock In \emph{{ICLR}}. OpenReview.net.

\bibitem[{Liu et~al.(2019)Liu, Ott, Goyal, Du, Joshi, Chen, Levy, Lewis, Zettlemoyer, and Stoyanov}]{RoBERTa}
Yinhan Liu, Myle Ott, Naman Goyal, Jingfei Du, Mandar Joshi, Danqi Chen, Omer Levy, Mike Lewis, Luke Zettlemoyer, and Veselin Stoyanov. 2019.
\newblock \href {https://doi.org/10.48550/arXiv.1907.11692} {Roberta: {A} robustly optimized {BERT} pretraining approach}.
\newblock \emph{CoRR}, abs/1907.11692.

\bibitem[{Liu et~al.(2023{\natexlab{a}})Liu, Li, Luo, Fei, Cao, Kawaguchi, Wang, and Chua}]{MolCA}
Zhiyuan Liu, Sihang Li, Yanchen Luo, Hao Fei, Yixin Cao, Kenji Kawaguchi, Xiang Wang, and Tat{-}Seng Chua. 2023{\natexlab{a}}.
\newblock {M}ol{CA}: Molecular graph-language modeling with cross-modal projector and uni-modal adapter.
\newblock In \emph{{EMNLP}}, pages 15623--15638. Association for Computational Linguistics.

\bibitem[{Liu et~al.(2024{\natexlab{a}})Liu, Shi, Zhang, Li, Zhang, Wang, Kawaguchi, and Chua}]{ReactXT}
Zhiyuan Liu, Yaorui Shi, An~Zhang, Sihang Li, Enzhi Zhang, Xiang Wang, Kenji Kawaguchi, and Tat{-}Seng Chua. 2024{\natexlab{a}}.
\newblock Reactxt: Understanding molecular "reaction-ship" via reaction-contextualized molecule-text pretraining.
\newblock In \emph{{ACL} (Findings)}, pages 5353--5377. Association for Computational Linguistics.

\bibitem[{Liu et~al.(2023{\natexlab{b}})Liu, Shi, Zhang, Zhang, Kawaguchi, Wang, and Chua}]{SimSGT}
Zhiyuan Liu, Yaorui Shi, An~Zhang, Enzhi Zhang, Kenji Kawaguchi, Xiang Wang, and Tat-Seng Chua. 2023{\natexlab{b}}.
\newblock \href {https://openreview.net/forum?id=fWLf8DV0fI} {Rethinking tokenizer and decoder in masked graph modeling for molecules}.
\newblock In \emph{{NeurIPS}}.

\bibitem[{Liu et~al.(2024{\natexlab{b}})Liu, Zhang, Fei, Zhang, Wang, Kawaguchi, and Chua}]{ProtT3}
Zhiyuan Liu, An~Zhang, Hao Fei, Enzhi Zhang, Xiang Wang, Kenji Kawaguchi, and Tat-Seng Chua. 2024{\natexlab{b}}.
\newblock {P}rot{T}3: Protein-to-text generation for text-based protein understanding.
\newblock In \emph{{ACL}}, pages 5949--5966. Association for Computational Linguistics.

\bibitem[{Liu et~al.(2023{\natexlab{c}})Liu, Zhang, Sun, Li, Shi, Li, Wang, He, and Chua}]{E-GCL}
Zhiyuan Liu, An~Zhang, Yu~Sun, Yicong Li, Yaorui Shi, Sihang Li, Xiang Wang, Xiangnan He, and Tat-Seng Chua. 2023{\natexlab{c}}.
\newblock \href {https://openreview.net/forum?id=9L1Ts8t66YK} {Towards equivariant graph contrastive learning via cross-graph augmentation}.

\bibitem[{Luo et~al.(2021{\natexlab{a}})Luo, Shi, Xu, and Tang}]{DGSM}
Shitong Luo, Chence Shi, Minkai Xu, and Jian Tang. 2021{\natexlab{a}}.
\newblock \href {https://doi.org/10.5555/3540261.3541774} {Predicting molecular conformation via dynamic graph score matching}.
\newblock In \emph{NeurIPS}, pages 19784--19795.

\bibitem[{Luo et~al.(2021{\natexlab{b}})Luo, Yan, and Ji}]{GraphDF}
Youzhi Luo, Keqiang Yan, and Shuiwang Ji. 2021{\natexlab{b}}.
\newblock \href {https://doi.org/10.48550/arXiv.2102.01189} {Graphdf: {A} discrete flow model for molecular graph generation}.
\newblock In \emph{{ICML}}, volume 139 of \emph{Proceedings of Machine Learning Research}, pages 7192--7203. {PMLR}.

\bibitem[{Madhawa et~al.(2019)Madhawa, Ishiguro, Nakago, and Abe}]{GraphNVP}
Kaushalya Madhawa, Katushiko Ishiguro, Kosuke Nakago, and Motoki Abe. 2019.
\newblock \href {https://doi.org/10.48550/arXiv.1905.11600} {Graphnvp: An invertible flow model for generating molecular graphs}.
\newblock \emph{CoRR}, abs/1905.11600.

\bibitem[{Mandal et~al.(2009)Mandal, Moudgil, and Mandal}]{Mandal_Moudgil_Mandal_2009}
Soma Mandal, Mee'nal Moudgil, and Sanat~K. Mandal. 2009.
\newblock \href {https://doi.org/https://doi.org/10.1016/j.ejphar.2009.06.065} {Rational drug design}.
\newblock \emph{European Journal of Pharmacology}, 625(1):90--100.
\newblock New Vistas in Anti-Cancer Therapy.

\bibitem[{Mansimov et~al.(2019)Mansimov, Mahmood, Kang, and Cho}]{CVGAE}
Elman Mansimov, Omar Mahmood, Seokho Kang, and Kyunghyun Cho. 2019.
\newblock \href {https://doi.org/10.1038/s41598-019-56773-5} {Molecular geometry prediction using a deep generative graph neural network}.
\newblock \emph{Science Reports}, 9:20381.

\bibitem[{Nesterov et~al.(2020)Nesterov, Wieser, and Roth}]{3DMolNet}
Vitali Nesterov, Mario Wieser, and Volker Roth. 2020.
\newblock \href {https://doi.org/10.48550/arXiv.2010.06477} {3dmolnet: {A} generative network for molecular structures}.
\newblock \emph{CoRR}, abs/2010.06477.

\bibitem[{OpenAI(2023)}]{GPT-4}
OpenAI. 2023.
\newblock \href {https://doi.org/10.48550/arXiv.2303.08774} {{GPT-4} technical report}.
\newblock \emph{CoRR}, abs/2303.08774.

\bibitem[{Popova et~al.(2019)Popova, Shvets, Oliva, and Isayev}]{MolecularRNN}
Mariya Popova, Mykhailo Shvets, Junier Oliva, and Olexandr Isayev. 2019.
\newblock \href {https://doi.org/10.48550/arXiv.1905.13372} {Molecularrnn: Generating realistic molecular graphs with optimized properties}.
\newblock \emph{CoRR}, abs/1905.13372.

\bibitem[{Pyzer-Knapp et~al.(2015)Pyzer-Knapp, Suh, Gómez-Bombarelli, Aguilera-Iparraguirre, and Aspuru-Guzik}]{Pyzer-Knapp_2015}
Edward~O. Pyzer-Knapp, Changwon Suh, Rafael Gómez-Bombarelli, Jorge Aguilera-Iparraguirre, and Alán Aspuru-Guzik. 2015.
\newblock \href {https://doi.org/10.1146/annurev-matsci-070214-020823} {What is high-throughput virtual screening? a perspective from organic materials discovery}.
\newblock \emph{Annual Review of Materials Research}, 45:195–216.

\bibitem[{Raffel et~al.(2020)Raffel, Shazeer, Roberts, Lee, Narang, Matena, Zhou, Li, and Liu}]{T5}
Colin Raffel, Noam Shazeer, Adam Roberts, Katherine Lee, Sharan Narang, Michael Matena, Yanqi Zhou, Wei Li, and Peter~J. Liu. 2020.
\newblock \href {https://doi.org/10.5555/3455716.3455856} {Exploring the limits of transfer learning with a unified text-to-text transformer}.
\newblock \emph{Journal of Machine Learning Research}, 21:140:1--140:67.

\bibitem[{Ramakrishnan et~al.(2014)Ramakrishnan, Dral, Rupp, and von Lilienfeld}]{QM9}
Raghunathan Ramakrishnan, Pavlo~O. Dral, Matthias Rupp, and O.~Anatole von Lilienfeld. 2014.
\newblock \href {https://doi.org/10.1038/sdata.2014.22} {Quantum chemistry structures and properties of 134 kilo molecules}.
\newblock \emph{Scientific Data}, 1:140022.

\bibitem[{Rombach et~al.(2022)Rombach, Blattmann, Lorenz, Esser, and Ommer}]{LDMs}
Robin Rombach, Andreas Blattmann, Dominik Lorenz, Patrick Esser, and Bj{\"{o}}rn Ommer. 2022.
\newblock \href {https://doi.org/10.48550/arXiv.2112.10752} {High-resolution image synthesis with latent diffusion models}.
\newblock In \emph{{CVPR}}, pages 10674--10685. {IEEE}.

\bibitem[{Ruiz et~al.(2023)Ruiz, Li, Jampani, Pritch, Rubinstein, and Aberman}]{DreamBooth}
Nataniel Ruiz, Yuanzhen Li, Varun Jampani, Yael Pritch, Michael Rubinstein, and Kfir Aberman. 2023.
\newblock \href {https://doi.org/10.48550/arXiv.2208.12242} {Dreambooth: Fine tuning text-to-image diffusion models for subject-driven generation}.
\newblock In \emph{{CVPR}}, pages 22500--22510. {IEEE}.

\bibitem[{Rutz et~al.(2022)Rutz, Sorokina, Galgonek, Mietchen, Willighagen, Gaudry, Graham, Stephan, Page, Vondrášek, Steinbeck, Pauli, Wolfender, Bisson, and Allard}]{LOTUS}
Adriano Rutz, Maria Sorokina, Jakub Galgonek, Daniel Mietchen, Egon Willighagen, Arnaud Gaudry, James~G Graham, Ralf Stephan, Roderic Page, Jiří Vondrášek, Christoph Steinbeck, Guido~F Pauli, Jean-Luc Wolfender, Jonathan Bisson, and Pierre-Marie Allard. 2022.
\newblock \href {https://doi.org/10.7554/eLife.70780} {The lotus initiative for open knowledge management in natural products research}.
\newblock \emph{eLife}, 11:e70780.

\bibitem[{Saharia et~al.(2023)Saharia, Ho, Chan, Salimans, Fleet, and Norouzi}]{SR3}
Chitwan Saharia, Jonathan Ho, William Chan, Tim Salimans, David~J. Fleet, and Mohammad Norouzi. 2023.
\newblock \href {https://doi.org/10.1109/TPAMI.2022.3204461} {Image super-resolution via iterative refinement}.
\newblock \emph{IEEE Transactions on Pattern Analysis and Machine Intelligence}, 45(4):4713--4726.

\bibitem[{Sanchez-Lengeling and Aspuru-Guzik(2018)}]{Inverse_molecular_design}
Benjamin Sanchez-Lengeling and Alán Aspuru-Guzik. 2018.
\newblock \href {https://doi.org/10.1126/science.aat2663} {Inverse molecular design using machine learning: Generative models for matter engineering}.
\newblock \emph{Science}, 361(6400):360--365.

\bibitem[{Satorras et~al.(2021{\natexlab{a}})Satorras, Hoogeboom, Fuchs, Posner, and Welling}]{E-NFs}
Victor~Garcia Satorras, Emiel Hoogeboom, Fabian Fuchs, Ingmar Posner, and Max Welling. 2021{\natexlab{a}}.
\newblock E(n) equivariant normalizing flows.
\newblock In \emph{NeurIPS}, pages 4181--4192.

\bibitem[{Satorras et~al.(2021{\natexlab{b}})Satorras, Hoogeboom, and Welling}]{EGNN}
Victor~Garcia Satorras, Emiel Hoogeboom, and Max Welling. 2021{\natexlab{b}}.
\newblock \href {https://doi.org/10.48550/arXiv.2102.09844} {E(n) equivariant graph neural networks}.
\newblock In \emph{{ICML}}, volume 139 of \emph{Proceedings of Machine Learning Research}, pages 9323--9332. {PMLR}.

\bibitem[{Schneider(2023)}]{ArchiSound}
Flavio Schneider. 2023.
\newblock \href {https://doi.org/10.48550/ARXIV.2301.13267} {Archisound: Audio generation with diffusion}.
\newblock \emph{CoRR}, abs/2301.13267.

\bibitem[{Shi et~al.(2020)Shi, Xu, Zhu, Zhang, Zhang, and Tang}]{GraphAF}
Chence Shi, Minkai Xu, Zhaocheng Zhu, Weinan Zhang, Ming Zhang, and Jian Tang. 2020.
\newblock \href {https://doi.org/10.48550/arXiv.2001.09382} {Graphaf: a flow-based autoregressive model for molecular graph generation}.
\newblock In \emph{{ICLR}}. OpenReview.net.

\bibitem[{Simonovsky and Komodakis(2018)}]{GraphVAE}
Martin Simonovsky and Nikos Komodakis. 2018.
\newblock \href {https://doi.org/10.48550/arXiv.1802.03480} {Graphvae: Towards generation of small graphs using variational autoencoders}.
\newblock In \emph{{ICANN} {(1)}}, volume 11139 of \emph{Lecture Notes in Computer Science}, pages 412--422. Springer.

\bibitem[{Sohl{-}Dickstein et~al.(2015)Sohl{-}Dickstein, Weiss, Maheswaranathan, and Ganguli}]{Diffusion}
Jascha Sohl{-}Dickstein, Eric~A. Weiss, Niru Maheswaranathan, and Surya Ganguli. 2015.
\newblock \href {https://doi.org/10.48550/arXiv.1503.03585} {Deep unsupervised learning using nonequilibrium thermodynamics}.
\newblock In \emph{{ICML}}, volume~37 of \emph{{JMLR} Workshop and Conference Proceedings}, pages 2256--2265. JMLR.org.

\bibitem[{Song et~al.(2021)Song, Sohl{-}Dickstein, Kingma, Kumar, Ermon, and Poole}]{SDE}
Yang Song, Jascha Sohl{-}Dickstein, Diederik~P. Kingma, Abhishek Kumar, Stefano Ermon, and Ben Poole. 2021.
\newblock \href {https://doi.org/10.48550/arXiv.2011.13456} {Score-based generative modeling through stochastic differential equations}.
\newblock In \emph{{ICLR}}. OpenReview.net.

\bibitem[{Su et~al.(2022)Su, Du, Yang, Zhou, Li, Rao, Sun, Lu, and Wen}]{MoMu}
Bing Su, Dazhao Du, Zhao Yang, Yujie Zhou, Jiangmeng Li, Anyi Rao, Hao Sun, Zhiwu Lu, and Ji{-}Rong Wen. 2022.
\newblock \href {https://doi.org/10.48550/ARXIV.2209.05481} {A molecular multimodal foundation model associating molecule graphs with natural language}.
\newblock \emph{CoRR}, abs/2209.05481.

\bibitem[{Thomas et~al.(2018)Thomas, Smidt, Kearnes, Yang, Li, Kohlhoff, and Riley}]{TFN}
Nathaniel Thomas, Tess~E. Smidt, Steven Kearnes, Lusann Yang, Li~Li, Kai Kohlhoff, and Patrick Riley. 2018.
\newblock \href {https://doi.org/10.48550/arXiv.1802.08219} {Tensor field networks: Rotation- and translation-equivariant neural networks for 3d point clouds}.
\newblock \emph{CoRR}, abs/1802.08219.

\bibitem[{Weininger(1988)}]{SMILES}
David Weininger. 1988.
\newblock \href {https://doi.org/10.1021/CI00057A005} {Smiles, a chemical language and information system. 1. introduction to methodology and encoding rules}.
\newblock \emph{Journal of Chemical Information and Computer Sciences}, 28(1):31--36.

\bibitem[{Willmott and Matsuura(2005)}]{MAE}
Cort~J Willmott and Kenji Matsuura. 2005.
\newblock \href {https://doi.org/10.3354/cr030079} {Advantages of the mean absolute error (mae) over the root mean square error (rmse) in assessing average model performance}.
\newblock \emph{Climate research}, 30(1):79--82.

\bibitem[{Wishart et~al.(2014)Wishart, Arndt, Pon, Sajed, Guo, Djoumbou, Knox, Wilson, Liang, Grant, Liu, Goldansaz, and Rappaport}]{T3DB}
David Wishart, David Arndt, Allison Pon, Tanvir Sajed, An~Chi Guo, Yannick Djoumbou, Craig Knox, Michael Wilson, Yongjie Liang, Jason Grant, Yifeng Liu, Seyed Goldansaz, and Stephen Rappaport. 2014.
\newblock \href {https://doi.org/10.1093/nar/gku1004} {T3db: The toxic exposome database}.
\newblock \emph{Nucleic acids research}, 43.

\bibitem[{Wu et~al.(2022)Wu, Gong, Liu, Ye, and Liu}]{Bridge}
Lemeng Wu, Chengyue Gong, Xingchao Liu, Mao Ye, and Qiang Liu. 2022.
\newblock \href {https://doi.org/10.48550/arXiv.2209.00865} {Diffusion-based molecule generation with informative prior bridges}.
\newblock In \emph{NeurIPS}.

\bibitem[{Xu et~al.(2019)Xu, Hu, Leskovec, and Jegelka}]{GIN}
Keyulu Xu, Weihua Hu, Jure Leskovec, and Stefanie Jegelka. 2019.
\newblock \href {https://doi.org/10.48550/arXiv.1810.00826} {How powerful are graph neural networks?}
\newblock In \emph{{ICLR}}.

\bibitem[{Xu et~al.(2022)Xu, Yu, Song, Shi, Ermon, and Tang}]{GeoDiff}
Minkai Xu, Lantao Yu, Yang Song, Chence Shi, Stefano Ermon, and Jian Tang. 2022.
\newblock \href {https://doi.org/10.48550/arXiv.2203.02923} {Geodiff: {A} geometric diffusion model for molecular conformation generation}.
\newblock In \emph{{ICLR}}. OpenReview.net.

\bibitem[{Yang et~al.(2023)Yang, Sun, Liu, Xue, Deng, and Wang}]{CMGN}
Minjian Yang, Hanyu Sun, Xue Liu, Xi~Xue, Yafeng Deng, and Xiaojian Wang. 2023.
\newblock \href {https://doi.org/10.1093/bib/bbad185} {{CMGN:} a conditional molecular generation net to design target-specific molecules with desired properties}.
\newblock \emph{Briefings Bioinform.}, 24(4).

\bibitem[{Zang and Wang(2020)}]{MoFlow}
Chengxi Zang and Fei Wang. 2020.
\newblock \href {https://doi.org/10.1145/3394486.3403104} {Moflow: An invertible flow model for generating molecular graphs}.
\newblock In \emph{{KDD}}, pages 617--626. {ACM}.

\bibitem[{Zeng et~al.(2022)Zeng, Yao, Liu, and Sun}]{KVPLM}
Zheni Zeng, Yuan Yao, Zhiyuan Liu, and Maosong Sun. 2022.
\newblock \href {https://doi.org/10.1038/s41467-022-28494-3} {A deep-learning system bridging molecule structure and biomedical text with comprehension comparable to human professionals}.
\newblock \emph{Nature communications}, 13(862).

\end{thebibliography}
    \bibliographystyle{acl_natbib}

	\else
		\clearpage
		\section*{Supplementary Information}

		\clearpage

	\fi

\end{document}